\documentclass[lettersize,journal]{IEEEtran}
\usepackage{amsmath,amsfonts}
\usepackage{algorithm}
\usepackage{lineno}
\usepackage{array}
\usepackage[caption=false,font=normalsize,labelfont=sf,textfont=sf]{subfig}
\usepackage{textcomp}
\usepackage{stfloats}
\usepackage{url}
\usepackage{verbatim}
\usepackage{graphicx}
\usepackage{cite}
\usepackage{marvosym}

\usepackage{adjustbox}

\usepackage{utfsym}

\usepackage[table]{xcolor}
\usepackage{multirow,algorithm}
\usepackage[utf8]{inputenc} 
\usepackage[T1]{fontenc}    
\usepackage{booktabs}       
\usepackage{amsfonts}       
\usepackage{nicefrac}       
\usepackage{microtype}      
\usepackage[dvipsnames]{xcolor}
\usepackage{colortbl}
\usepackage{enumitem}
\usepackage{algpseudocode}
\usepackage{times}
\usepackage{epsfig}
\usepackage{amssymb}
\usepackage{caption}
\usepackage{subcaption}
\usepackage{tabularx}
\usepackage{rotating}
\usepackage{etoolbox}
\usepackage[normalem]{ulem}
\usepackage{lipsum}
\usepackage{stmaryrd}
\usepackage{stackengine}
\usepackage{makecell}
\usepackage{pifont}
\usepackage{cancel}
\usepackage{wrapfig}
\usepackage{bm}

\usepackage{tikz}

\definecolor{noise}{RGB}{255, 255, 255} 
\definecolor{other-ground}{RGB}{112,180,60} 
\definecolor{vehicle}{RGB}{255, 158, 0} 
\definecolor{bicycle}{RGB}{220,20,60} 
\definecolor{pedestrian}{RGB}{0,0,230} 
\definecolor{traffic-cone}{RGB}{47,79,79} 
\definecolor{barrier}{RGB}{112,128,144} 
\definecolor{construction-zones}{RGB}{255, 200, 0} 
\definecolor{generic-object}{RGB}{222,184,13}
\definecolor{road}{RGB}{0,207,191} 
\definecolor{road-line}{RGB}{150 , 240, 80}

\definecolor{y}{HTML}{00994C}
\definecolor{b}{rgb}{0.31372549, 0.11764706, 0.70588235F}

\def\eg{\textit{e.g.}}
\def\ie{\textit{i.e.}}

\def\ours{UniScenev2}

\newcommand{\re}[1]{\textcolor{black}{#1}}
\newcommand{\rcap}{\captionsetup{labelfont={color=black}, textfont={color=black}}}

\begin{document}

\title{ 
Scaling Up Occupancy-centric Driving Scene Generation: Dataset and Method
}

\author{Bohan Li,~Xin Jin,~Hu Zhu,~Hongsi Liu,~Ruikai Li,~Jiazhe Guo,~Kaiwen Cai,\\~Chao Ma,~Yueming Jin,~Hao Zhao,~Xiaokang Yang,~\IEEEmembership{Fellow, IEEE},~Wenjun Zeng\textsuperscript{\Letter},~\IEEEmembership{Fellow, IEEE}
\vspace{-7pt}
\thanks{Bohan Li is with Shanghai Jiao Tong University, Shanghai, China, and Eastern Institute of Technology, Ningbo, China. Xiaokang Yang is a distinguished professor, and Chao Ma is an associate professor at the School of Electronic Information and Electrical Engineering, Shanghai Jiao Tong University, Shanghai, China. Hu Zhu, and Hongsi Liu are with Eastern Institute of Technology, Ningbo, China.
}	
\thanks{Ruikai Li, Jiazhe Guo, Kaiwen Cai are with Li Auto, Beijing, China.}
\thanks{Yueming Jin is with National University of Singapore, Singapore.}
\thanks{Hao Zhao is with Tsinghua University, Beijing, China.}
\thanks{Xin Jin is an assistant professor, and Wenjun Zeng (corresponding author) is a chair professor at the Ningbo Key Laboratory of Spatial Intelligence and Digital Derivative, Ningbo Institute of Digital Twin, Eastern Institute of Technology, Ningbo, Zhejiang, China, (e-mail: wzeng-vp@eitech.edu.cn).}
}

\markboth{SUBMITTED TO IEEE Transactions on Pattern Analysis and Machine Intelligence}%
{Shell \MakeLowercase{\textit{et al.}}: A Sample Article Using IEEEtran.cls for IEEE Journals}


\maketitle

\begin{abstract}  
Driving scene generation is a critical domain for autonomous driving, enabling downstream applications, including perception and planning evaluation. Occupancy-centric methods have recently achieved state-of-the-art results by offering consistent conditioning across frames and modalities; however, their performance heavily depends on annotated occupancy data, which still remains scarce. To overcome this limitation, we curate Nuplan-Occ, the largest semantic occupancy dataset to date, constructed from the widely used Nuplan benchmark. 
Its scale and diversity facilitate not only large-scale generative modeling but also autonomous driving downstream applications. 
Based on this dataset, we develop a unified framework that jointly synthesizes high-quality semantic occupancy, multi-view videos, and LiDAR point clouds. Our approach incorporates a spatio-temporal disentangled architecture to support high-fidelity spatial expansion and temporal forecasting of 4D dynamic occupancy. To bridge modal gaps, we further propose two novel techniques: a Gaussian splatting-based sparse point map rendering strategy that enhances multi-view video generation, and a sensor-aware embedding strategy that explicitly models LiDAR sensor properties for realistic multi-LiDAR simulation. Extensive experiments demonstrate that our method achieves superior generation fidelity and scalability compared to existing approaches, and validates its practical value in downstream tasks.\looseness=-1
\end{abstract}


\section{Introduction}

The generation of high-quality driving scenes represents a promising avenue for advancing autonomous driving (AD), as it alleviates the significant resource demands associated with real-world data collection and annotation~\cite{luo2021diffusion, rombach2022high, jiang2022conditional,li2024time,mao2024dreamdrive,li2024uniscene,wang2024stag,song2025history}. Recent breakthroughs in generative models, particularly diffusion-based approaches~\cite{luo2021diffusion, rombach2022high, jiang2022conditional, li2024time}, have enabled the creation of highly realistic synthetic data~\cite{yang2023bevcontrol, swerdlow2024street, wang2023drivedreamer}, thereby facilitating advancements in downstream tasks. 
Current driving scene generation works~\cite{wang2023drivedreamer, gao2023magicdrive,wen2023panacea, zyrianov2024lidardm} commonly rely on layout conditions derived from coarse geometric labels, such as bird's-eye-view (BEV) maps and 3D bounding boxes, to guide the scene generation process. The synthetic data produced through these methods is subsequently utilized to enhance the performance of downstream tasks, including BEV segmentation~\cite{li2023open, wu2023datasetdm, li2024fairdiff} and 3D object detection~\cite{bowles2018gan, chen2023integrating, wang2024detdiffusion, he2022synthetic, moller2023prompt}.

These driving scene generation models predominantly focus on producing data in a single format (\eg, RGB video)~\cite{wang2023drivedreamer,zhao2024drivedreamer4d,gao2023magicdrive,wen2024panacea}, without fully leveraging the potential of generating data across multiple formats. This limitation restricts their applicability to a broad range of downstream tasks that rely on diverse sensor data, including RGB video and LiDAR point clouds in real-world scenarios~\cite{wang2023openoccupancy,liang2022bevfusion,li2022deepfusion,bai2022transfusion,berrio2021camera,zyrianov2022learning, ran2024towards,zyrianov2024lidardm, li2024time}. 
Furthermore, existing methods typically attempt to model the real-world distribution using a single-step layout-to-data generation process based solely on coarse input conditions (\eg, BEV layouts or 3D bounding boxes)~\cite{gao2023magicdrive,wen2024panacea,zhao2024drivedreamer2}. This direct learning approach often undermines the model's ability to capture the intricate distributions inherent in real-world driving scenes (\eg, realistic geometry and appearance) and leads to suboptimal performance.\looseness=-1

\begin{figure*}[!ht]
\centering
\includegraphics[width=0.85\linewidth]{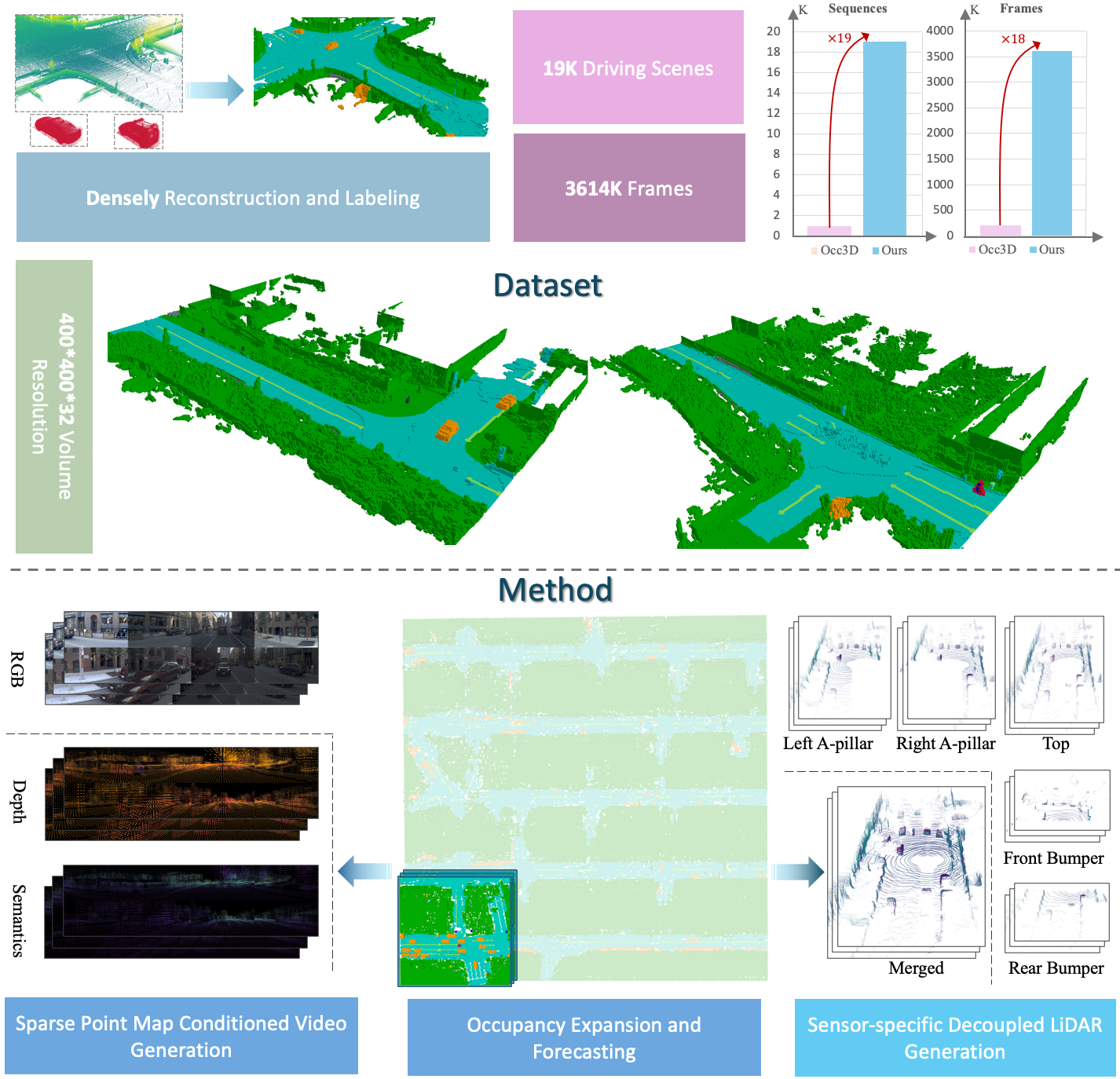}
		\begin{tabular}{ccccccccc}	
			\multicolumn{9}{c}{
				\small
                \textcolor{other-ground}{$\blacksquare$}other-ground
                \textcolor{vehicle}{$\blacksquare$}vehicle
                \textcolor{bicycle}{$\blacksquare$}bicycle
                \textcolor{pedestrian}{$\blacksquare$}pedestrian
                \textcolor{traffic-cone}{$\blacksquare$}traffic-cone
                \textcolor{barrier}{$\blacksquare$}barrier
                \textcolor{construction-zones}{$\blacksquare$}construction-zones
                \textcolor{generic-object}{$\blacksquare$}generic-object   
                \textcolor{road}{$\blacksquare$}road
                \textcolor{road-line}{$\blacksquare$}road-line
			}
		\end{tabular}
    \caption{{Overview of Nuplan-Occ dataset and the \ours\ pipeline.} We introduce the largest semantic occupancy dataset to date, featuring dense 3D semantic annotations that contain $\sim$19× more annotated scenes and $\sim$18× more frames than Nuscenes-Occupancy~\cite{tian2024occ3d,wei2023surroundocc}. Facilitated with Nuplan-Occ, \ours\ scales up both model architecture and training data to enable high-quality occupancy spatial expansion and temporal forecasting, as well as occupancy-based sparse point map condition for video generation, and sensor-specific LiDAR generation. }
\label{fig_teaser}
\vspace{-0pt}
\end{figure*}

To address these challenges, UniScene~\cite{li2024uniscene} proposes to utilize 3D semantic occupancy as an intermediate representation with rich semantic and geometric information to decompose complex autonomous driving scene generation tasks into hierarchical steps for high-quality multi-modal generation of semantic occupancy, video, and LiDAR data~\cite{tong2023scene, hu2023planning, wang2023openoccupancy, zyrianov2022learning, ran2024towards, zyrianov2024lidardm, li2024time}. 
Within the framework, 3D semantic occupancy is first generated from BEV scene layouts and then utilized to guide the subsequent generation of video and LiDAR data. The generated semantic occupancy serves as an intermediate representation, guiding the subsequent generation of other output modalities with 3D structural details and semantic priors.

However, the generation capabilities of UniScene remain constrained by limited scenario diversity and scale, akin to prior works~\cite{wang2023drivedreamer, gao2023magicdrive,wen2023panacea, zyrianov2024lidardm,zhao2024drivedreamer2,li2025uniscene}, which limits its practical utility for scalable downstream tasks.
To address these limitations, we propose \ours, a unified occupancy-centric framework for versatile 4D dynamic scene generation of semantic occupancy, video, and LiDAR data.
Beyond UniScene\cite{li2024uniscene}, which generates 3D semantic occupancy, multi-view video, and LiDAR data via a decomposed learning paradigm and hierarchical architecture, \ours\ overcomes its predecessor’s scalability constraints on the Nuscenes \cite{caesar2020nuscenes} dataset. By scaling both model architecture and training data, \ours\ achieves large-scale semantic occupancy generation and synthesizes corresponding multi-view videos and LiDAR point clouds, as shown in Figure~\ref{fig_teaser}.

Specifically, to enable efficient training across diverse autonomous driving scenarios, we construct Nuplan-Occ, a large-scale semantic occupancy dataset extending the Nuplan~\cite{Nuplan} benchmark with dense 3D semantic annotations. As detailed in Figure~\ref{fig_teaser} and Table~\ref{tab_dataset}, Nuplan-Occ contains $\sim$19× more annotated scenes and $\sim$18× more frames than Nuscenes-Occupancy~\cite{tian2024occ3d,wei2023surroundocc}. Our automated annotation pipeline employs a Foreground-Background Separate Aggregation Strategy (FBSA) for dense reconstruction and precise semantic label assignment, which reconstructs dense semantic occupancy grids by separately aggregating foreground objects and background content from multi-frame LiDAR scans. This process involves point-based registration, denoising, neural kernel-based reconstruction~\cite{huang2023nksr}, and voxelization for precise semantic labeling.

\begin{table*}[!ht]
\vspace{-0pt}
\begin{center}
\scriptsize
\vspace{-0pt}
\renewcommand\tabcolsep{3.5pt}
\centering
\resizebox{0.95\linewidth}{!}{
\begin{tabular}{l|cccccccc}
\toprule Dataset & Type & Surrounded & View & Modility & \#Sequence  & \#Frames & Volume Size & Resolution(m) \\ \midrule
NNYUv2~\cite{silberman2012indoor}   & Indoor  & \textcolor{red}{\usym{2717}} &1 & C\&D &0.5K  & 1.4K & [240,240,14] & - \\ 

SceneNN~\cite{hua2016scenenn} & Indoor  & \textcolor{red}{\usym{2717}} & 1 & C\&D  & 100 & - & - &  -\\
SynthCity~\cite{griffiths2019synthcity} & Indoor  & \textcolor{red}{\usym{2717}} & 1 & C\&D  & 9 & - & -  & - \\
ScanNet~\cite{dai2017scannet}  & Indoor  & \textcolor{red}{\usym{2717}} &1 & C\&D & 1.5K & 1.5K & [62,62,31] & - \\
SemanticPOSS~\cite{pan2020semanticposs} & Indoor  & \textcolor{ForestGreen}{\usym{2713}} &1  & C\&D  &- &3K &- & - \\
SemanticKITTI~\cite{behley2019semantickitti} & Outdoor  & \textcolor{red}{\usym{2717}} &2 & C\&L & 22 & 4K & [256,256,32] & [0.2,0.2,0.2]  \\
KITTI-360~\cite{liao2022kitti360} & Outdoor & Fisheye &2 & C\&L & 11 & 90K & [256,256,32] &  [0.2,0.2,0.2] \\
SurroundOcc~\cite{wei2023surroundocc} & Outdoor & \textcolor{ForestGreen}{\usym{2713}}  &6 & C\&L & 1K & 200K & [200,200,16] & [0.5,0.5,0.5] \\  
Occ3D~\cite{tian2024occ3d} & Outdoor & \textcolor{ForestGreen}{\usym{2713}} &6 & C\&L & 1K & 200K & [200,200,16] & [0.4,0.4,0.4] \\  
OpenScene~\cite{openscene2023} & Outdoor & \textcolor{ForestGreen}{\usym{2713}} & 8 & C\&L & 1.8K & 69K & [200,200,16] & [0.5,0.5,0.5]   \\
\midrule
{Nuplan-Occ} (Ours) & Outdoor  &\textcolor{ForestGreen}{\usym{2713}} &8 & C\&L & 19K & 3614K & [400,400,32] & [0.25,0.25,0.25] \\

\bottomrule
\end{tabular}
 }
 \vspace{-0pt}
\caption{ {Comparison between Nuplan-Occ and other occupancy/LiDAR datasets.} ``Surrounded'' represents surround-view image inputs. ``View'' means the number of image view inputs. ``C'', ``D'', and ``L'' denote camera, depth, and LiDAR, respectively. }
\vspace{-0pt}
\label{tab_dataset}
\end{center}
\end{table*}

\begin{figure*}[!ht]
   \vspace{-0pt}
	\begin{center}
		\includegraphics[width=0.99\linewidth]{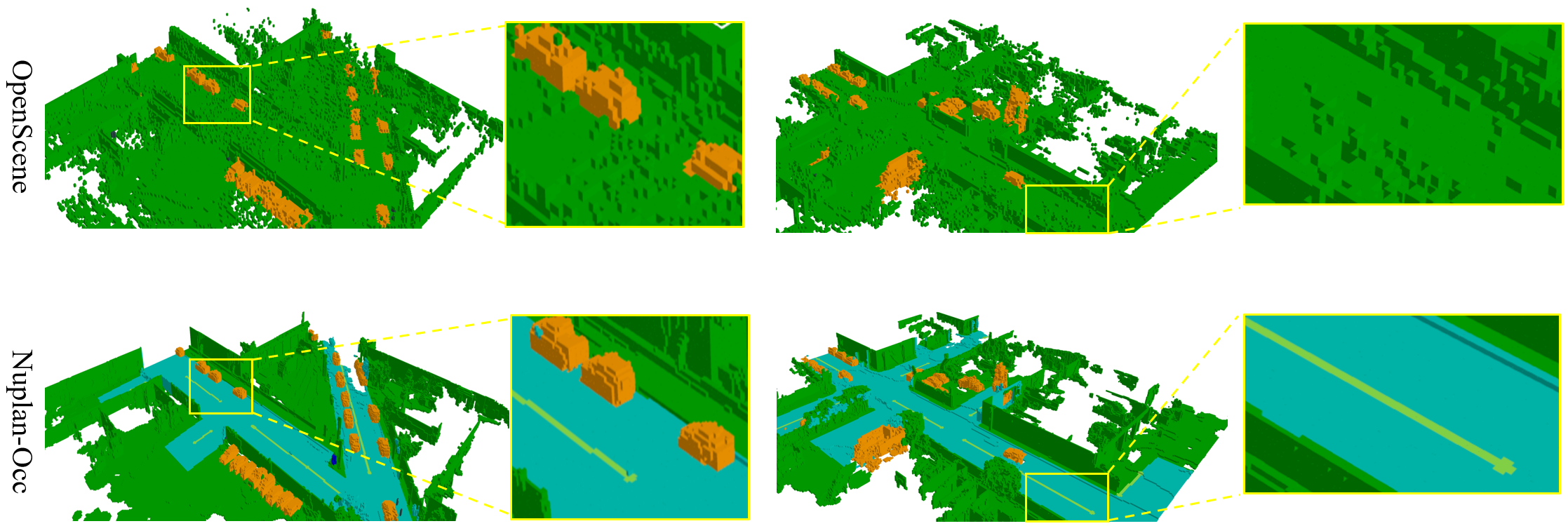}   
        \vspace{0pt}
		\begin{tabular}{ccccccccc}	
			\multicolumn{9}{c}{
				\small
                \textcolor{other-ground}{$\blacksquare$}other-ground
                \textcolor{vehicle}{$\blacksquare$}vehicle
                \textcolor{bicycle}{$\blacksquare$}bicycle
                \textcolor{pedestrian}{$\blacksquare$}pedestrian
                \textcolor{traffic-cone}{$\blacksquare$}traffic-cone
                \textcolor{barrier}{$\blacksquare$}barrier
                \textcolor{construction-zones}{$\blacksquare$}construction-zones
                \textcolor{generic-object}{$\blacksquare$}generic-object   
                \textcolor{road}{$\blacksquare$}road
                \textcolor{road-line}{$\blacksquare$}road-line
			}
		\end{tabular}	
 	\end{center}
        \vspace{-0pt}
	\caption{ The Nuplan-Occ provides dense semantic occupancy labels for 10HZ all frames in the Nuplan~\cite{Nuplan} dataset. Compared with OpenScene~\cite{openscene2023}, our method demonstrates high resolution (400$\times$400$\times$32) dense annotations with accurate geometry (\eg, clear vehicle structures and smooth road surfaces).
 } 
	\label{fig_open}
 \vspace{-0pt}
\end{figure*}

For framework design, a spatio-temporal disentangled architecture is introduced to enable high-quality spatial expansion and temporal forecasting for large-scale 4D
occupancy generation.
To bridge the representation gap and ensure high-quality, robust generation of video and LiDAR data on the large-scale Nuplan-Occ dataset, we introduce two novel modality-specific representation transfer strategies.
As shown in Figure~\ref{fig_teaser}, for multi-view video, a Gaussian splatting-based sparse point map rendering method provides robust conditional guidance, mitigating sensor calibration misalignment and noise in large-scale Nuplan~\cite{Nuplan} data. 
For LiDAR point clouds, a sensor-specific embedding strategy leveraging sensor position and ray information is proposed to explicitly simulate different LiDAR patterns.
This work extends the previous CVPR-25 conference paper UniScene~\cite{li2024uniscene} with substantial methodological advances, dataset innovation, and performance improvements. The key new contributions are:

1) A scalable framework for unified 4D dynamic scene generation.  
UniScenev2 overcomes the scale limitations of its predecessor by jointly scaling model architecture and training data. Built upon Nuplan-Occ—the largest semantic occupancy dataset to date, with $\sim$19× more scenes and $\sim$18× more frames, our approach achieves high-fidelity unified generation of semantic occupancy, multi-view video, and LiDAR data, leading to significant gains across tasks (\textit{e.g.}, 29.73\% mIoU in occupancy, 29.17\% FVD in video, and 54.25\% MMD in LiDAR generation).

2) Spatio-temporal disentangled modeling for 4D occupancy synthesis.
We introduce a novel generation framework that decouples 4D scene synthesis into two complementary tasks: spatial expansion and temporal forecasting. A dedicated data filtering strategy is proposed to isolate ego-motion and object-motion patterns, enabling robust and scalable dynamic occupancy generation.

3) Modality-bridging strategies for multi-sensor realism.
To bridge modality gaps in large-scale settings, we propose:
(i) A sparse point rendering strategy to facilitate geometrically precise and noise-robust conditioning for multi-view video;
(ii) A sensor-specific embedding scheme that explicitly encodes LiDAR extrinsics and ray geometry, enabling flexible and realistic multi-LiDAR simulation.
 
4) The Nuplan-Occ dataset: a large-scale semantic occupancy benchmark.
We curate and release Nuplan-Occ, comprising 3.6M frames with high-resolution voxel annotations (400×400×32). By leveraging a novel Foreground-Background Separate Aggregation pipeline, the dataset delivers dense 3D semantic labels with high geometric accuracy and label consistency.

Our code, demo video, and dataset are available at \url{https://arlo0o.github.io/uniscenev2/}.

\section{Related Work}

\subsection{Semantic Occupancy Representation}
Semantic occupancy has emerged as a key 3D representation for perception and generation \cite{cao2022monoscene,li2024bridging,huang2023tri,li2024one,li2024hierarchical,zheng2023occworld,li2024time,wang2024occsora,li2025occscene,li2026hierarchical,yang2025orv}. Existing perception methods include MonoScene \cite{cao2022monoscene} and FB-Occ \cite{li2023fb} for monocular and BEV feature learning, TPVFormer \cite{huang2023tri} with tri-perspective views, and SurroundOcc \cite{wei2023surroundocc} for multi-view fusion. VPD \cite{li2024time} further applies diffusion models to occupancy prediction.
\re{While early methods often struggled with geometric information loss and semantic confusion in complex multi-object regions, recent methods have introduced cross-modal guidance to improve robustness. {For instance, SG-SSC~\cite{liu20252d} successfully addresses these ambiguities by employing a 2D semantic-guided fusion strategy combined with a volume-guided semantic predictor to enhance boundary precision in dense scenes.}
{Furthermore, the rapid evolution of deep generative models has fundamentally transformed large-scale scene synthesis, as comprehensively reviewed in recent literature~\cite{wen20253d}.} Extending 3D generation to unbounded 4D urban environments introduces severe structural complexities. {To tackle this, CityDreamer4D~\cite{xie2025compositional} introduced a compositional paradigm that explicitly separates dynamic objects from static layouts using specialized neural fields.}}
SemCity \cite{lee2024semcity} uses triplane diffusion for static scenes, while PyramidOcc \cite{liu2023pyramid} employs pyramid discrete diffusion for large scales. Temporal modeling is addressed by OccWorld \cite{zheng2023occworld} for forecasting and OccLlama \cite{wei2024occllama} with semantic reasoning, though methods like OccSora \cite{wang2024occsora} still trail ground-truth performance. Other notable works include occupancy anticipation \cite{ramakrishnan2020occupancy}, TRELLIS \cite{xiang2024structured} for flexible outputs, Drive-OccWorld \cite{yang2025driving} for vision-centric forecasting, and DynamicCity \cite{bian2025dynamiccity} with hexplane-based VAEs.
A common limitation of the methods above is the neglect of spatiotemporal decoupling, hindering high-quality dynamic scene synthesis. Our approach overcomes this via a disentangled architecture and dedicated data filtering strategy, enabling high-fidelity 4D occupancy generation through separate spatial expansion and temporal forecasting.

\subsection{Driving Video Generation}
Recent advances in controllable video generation have improved simulation realism for autonomous driving~\cite{gao2023magicdrive,gao2025vista,li2024hierarchical,li2023bridging,wang2024driving,mao2024dreamdrive,li2024uniscene,song2025history,li2025omninwm}. Early frameworks like BEVGen~\cite{swerdlow2024street}, DriveDreamer~\cite{wang2023drivedreamer}, MagicDrive~\cite{gao2023magicdrive}, and Panacea~\cite{wen2023panacea} focused on temporal video synthesis, while later methods such as Drive-WM~\cite{wang2024driving} incorporated world models for enhanced coherence. Vista~\cite{gao2024vista} adapts Stable Video Diffusion (SVD)~\cite{blattmann2023stable} for single-view generation with action control. WoVoGen~\cite{lu2023wovogen} predicts future frames and occupancy from past data using learned feature compression~\cite{radford2021learning}. Other approaches include MagicDriveDiT~\cite{gao2024magicdrivedit} for scalability via DiT architectures, DreamDrive~\cite{mao2024dreamdrive} for 4D scenes with Gaussian representations.
However, these methods predominantly rely on single-step generation from coarse inputs, which limits their capacity to model complex real-world distributions. In this work, we employ a hierarchical strategy, generating occupancy as an intermediate representation to guide subsequent synthesis with robust sparse rendering maps for high-quality outputs.

\subsection{LiDAR Point Clouds Generation}
Current LiDAR generation methods~\cite{zyrianov2022learning,hahner2022lidar,zyrianov2024lidardm,ran2024towards} primarily use GANs or diffusion models. LiDM~\cite{hahner2022lidar} employs a VQVAE with range maps, improving geometric fidelity via curve-wise compression, patch-wise encoding, and point-wise supervision. LiDARGen~\cite{zyrianov2022learning} uses a score-based diffusion model on equirectangular images but is limited by its 2.5D representation for complex geometries. UltraLiDAR~\cite{xiong2023learning} voxelizes points into a bird’s-eye-view (BEV) and uses a VQVAE with a generative transformer, though the 2D BEV often loses fine-grained detail. Rendering-based approaches include NeRF-LiDAR~\cite{Zhang_Zhang2024}, which uses NeRF for synthesis, and GS-LiDAR~\cite{jiang2025gslidar}, which applies Gaussian Splatting for faster, superior dynamic reconstruction. However, these methods generally overlook sensor-specific information, restricting generation to fixed patterns. We propose a 3D occupancy-based pipeline with sensor-specific embeddings for position and ray data, to enable flexible and accurate LiDAR simulation.

\begin{figure*}[!ht]
    \centering
    \includegraphics[width=0.99\linewidth]{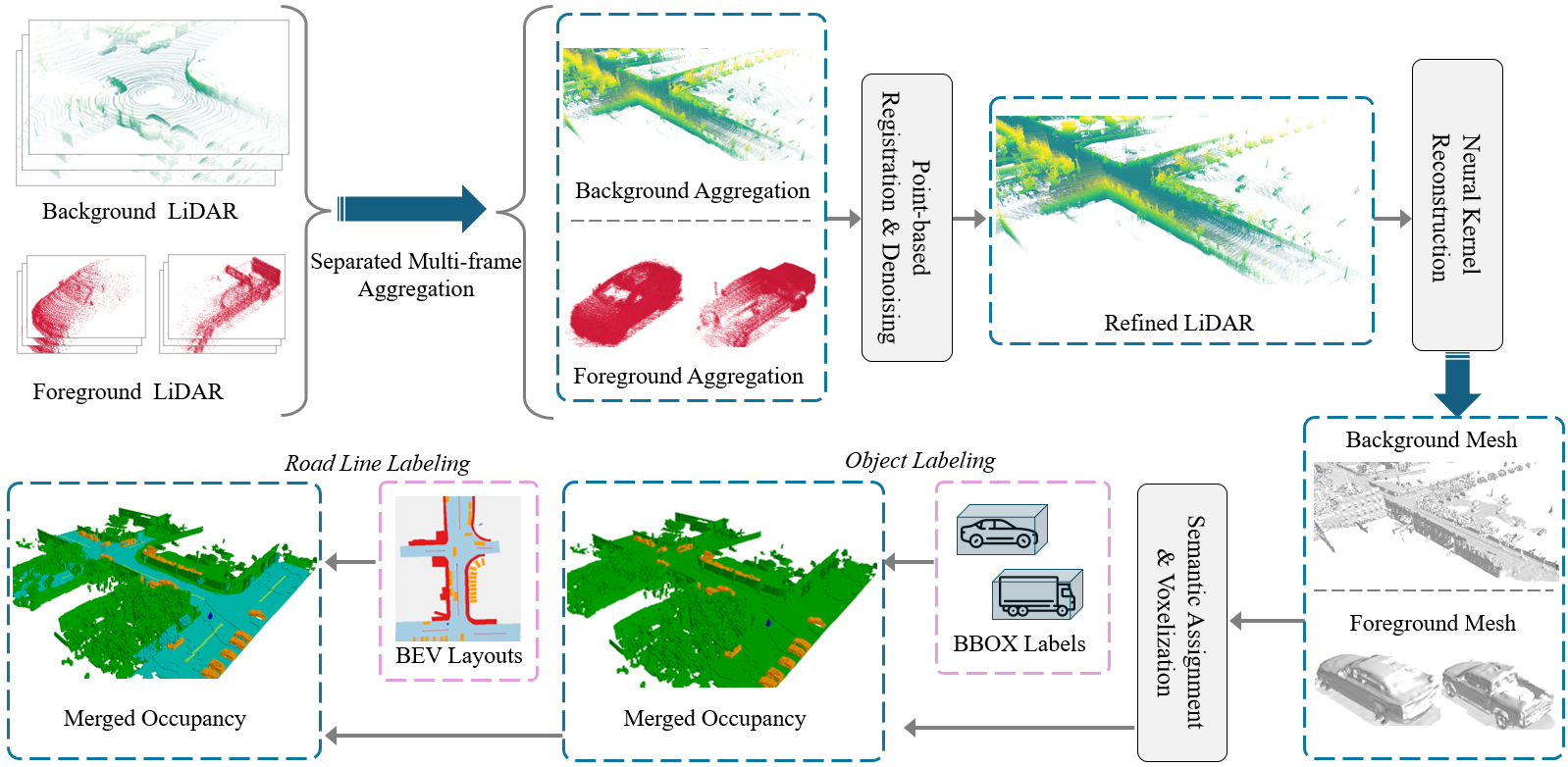}
        \begin{tabular}{ccccccccc}	
        \multicolumn{9}{c}{
            \small
            \textcolor{other-ground}{$\blacksquare$}other-ground
            \textcolor{vehicle}{$\blacksquare$}vehicle
            \textcolor{bicycle}{$\blacksquare$}bicycle
            \textcolor{pedestrian}{$\blacksquare$}pedestrian
            \textcolor{traffic-cone}{$\blacksquare$}traffic-cone
            \textcolor{barrier}{$\blacksquare$}barrier
            \textcolor{construction-zones}{$\blacksquare$}construction-zones
            \textcolor{generic-object}{$\blacksquare$}generic-object   
            \textcolor{road}{$\blacksquare$}road
            \textcolor{road-line}{$\blacksquare$}road-line
            }
        \end{tabular}	
        
    \caption{ {Nuplan-Occ dataset curation pipeline with the proposed Foreground-Background Separate Aggregation (FBSA) strategy.} 
     This strategy is composed of three key components: separated multi-frame point cloud aggregation, neural kernel-based mesh reconstruction, and hybrid semantic labeling.
    }
    \label{dataset-pipeline}
\end{figure*}

\section{Dataset}\label{sec_Nuplan}
This section introduces Nuplan-Occ, the largest semantic occupancy dataset to date, featuring dense 3D semantic annotations. 
To enhance reconstruction fidelity and label precision for data curation, we propose a Foreground-Background Separate Aggregation (FBSA) strategy. This strategy systematically addresses the challenges of occupancy densification and semantic label precision in the context of LiDAR-based 3D scene understanding. Below, we detail the methodology into three key components: separated point cloud aggregation, neural kernel-based mesh reconstruction, and hybrid semantic labeling.

\subsection{Separated Point Cloud Aggregation}
Given the sensor data from a scenario clip, we separate the point clouds of each frame into background points and object-specific foreground points based on object bounding boxes. The separation ensures that dynamic objects are treated independently from static environments, enabling more accurate processing for them.\looseness=-1

\noindent\textbf{Background Point Cloud Aggregation.}
First, we apply statistical filtering to the background points of each frame to reduce noise and prevent the occurrence of excessive floaters in the occupancy. The procedure can be described as follows:
\begin{equation}
    \mathbf{P}_\text{filtered} = \{ p \in \mathbf{P}_\text{refined} \mid \| p - \mu \| < k \cdot \sigma \},
\end{equation}
where $\mu$ and $\sigma$ represent the mean and standard deviation of the point cloud, respectively, and $k$ is a tunable threshold parameter.\looseness=-1 

Then, the background point clouds are aggregated into the world coordinate system using LiDAR extrinsics:
\begin{equation}
    \mathbf{P}_\text{world} = \mathbf{T}_\text{extrinsic} \cdot \mathbf{P}_\text{local},
\end{equation}
where $\mathbf{P}_\text{local}$ represents the local coordinates of the background points, and $\mathbf{T}_\text{extrinsic}$ denotes the transformation matrix encoding the LiDAR extrinsics. 
However, errors in $\mathbf{T}_\text{extrinsic}$ can degrade the quality of the aggregated point cloud, adversely affecting subsequent mesh reconstruction. 
To address this, we utilize Kiss-ICP~\cite{vizzo2023kissicp} for iterative point cloud registration, enabling explicit geometric alignment and refinement of the aggregated data.\looseness=-1

\noindent\textbf{Foreground Point Cloud Aggregation.}
For foreground object point clouds, we aggregate the points of each object into its local coordinate system based on its bounding box. Specifically, for an object with bounding box center $\mathbf{c}_\text{obj}$ and orientation $\mathbf{R}_\text{obj}$, the transformation to the local coordinate system is given by:\looseness=-1
\begin{equation}
    \mathbf{P}_\text{local}^\text{obj} = \mathbf{R}_\text{obj}^\top \cdot (\mathbf{P}_\text{world}^\text{obj} - \mathbf{c}_\text{obj}),
\end{equation}
where $\mathbf{P}_\text{world}^\text{obj}$ represents the foreground points in the world coordinate system.

\subsection{Neural Kernel-based Mesh Reconstruction}

To further increase point cloud density and improve surface representation, we use Neural Kernel Surface Reconstruction (NKSR)~\cite{huang2023nksr} to independently reconstruct meshes for both the aggregated background and each object point cloud. 

\noindent\textbf{Background Mesh Reconstruction.}
For the aggregated background point cloud $\mathbf{P}_\text{filtered}$, the reconstructed mesh vertices are extracted as densified points:
\begin{equation}
    \mathbf{V}_\text{bg} = \text{NKSR}(\mathbf{P}_\text{filtered}),
\end{equation}
where $\mathbf{V}_\text{bg}$ represents the vertices of the reconstructed background mesh.

\noindent\textbf{Foreground Mesh Reconstruction.}
Similarly, for each foreground object point cloud $\mathbf{P}_\text{local}^\text{obj}$, the corresponding mesh vertices are reconstructed as:
\begin{equation}
    \mathbf{V}_\text{fg}^\text{obj} = \text{NKSR}(\mathbf{P}_\text{local}^\text{obj}).
\end{equation}
These vertices are then transformed back to the world coordinate system for integration into the global scene.

\subsection{Hybrid Semantic Labeling}
The occupancy grids are extracted by voxelizing the reconstructed meshes, generating a compact 3D representation suitable for downstream tasks.
Semantic occupancy labels are derived by combining bounding box annotations for foreground objects and BEV map annotations for background regions. Since Nuplan does not provide point-level segmentation annotations, we adopt a hybrid approach for semantic labeling.\looseness=-1

\noindent\textbf{Foreground Object Labeling.}
Foreground objects are labeled using their bounding boxes. For a point $p \in \mathbf{P}_\text{world}$, the semantic label $l(p)$ is assigned as:
\begin{equation}
    l(p) = 
    \begin{cases} 
        \text{Object Class} & \text{if } p \in \text{BBox}(\mathbf{c}_\text{obj}, \mathbf{R}_\text{obj}), \\
        \text{Background} & \text{otherwise}.
    \end{cases}
\end{equation}

\noindent\textbf{Background Region Labeling.}
Background regions are labeled using the BEV map, which provides annotations for drivable areas and other semantic regions. For a voxel $v \in \mathbf{V}_\text{bg}$, the semantic label is determined by projecting the voxel with the BEV map:
\begin{equation}
    l(v) = \text{BEVLabel}(\text{Proj}(v)).
\end{equation}
where $\text{Proj}(v)$ denotes the projection of voxel $v$ with the correponding BEV map.\looseness=-1

\begin{figure*}[!ht]
    \centering
    \includegraphics[width=0.99\linewidth]{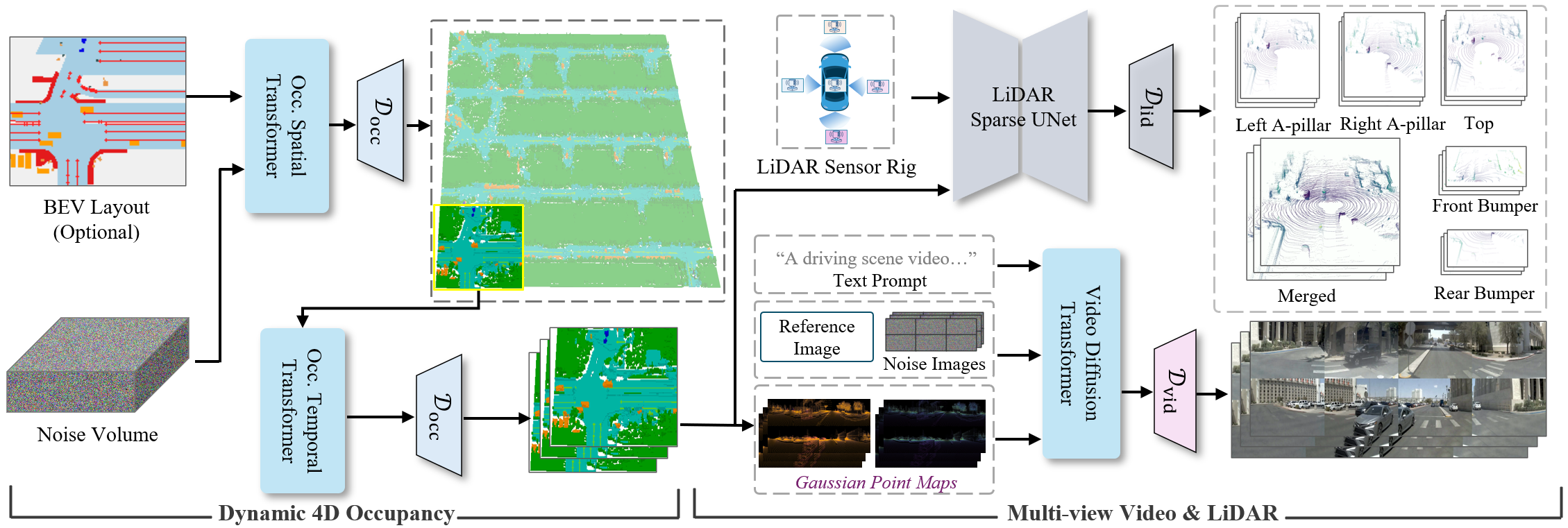}
    \caption{{Overall framework of \ours.}
     The joint generation process facilitates large-scale dynamic generation with an occupancy-centric hierarchy: {I. Dynamic Large-scale Occupancy Generation.} The optional BEV layout is concatenated with the noise volume before being fed into the occupancy spatial diffusion transformer, and decoded with the occupancy VAE decoder $\mathcal{D}_\mathrm{occ}$ to generate large-scale occupancy grids. 
     The occupancy temporal diffusion transformer processes a selected occupancy scene to forecast temporal occupancy sequences.
    {II. Occupancy-based Multi-view Video and LiDAR Generation.} 
     The occupancy is converted into 3D Gaussians and rendered into sparse semantic and depth point maps, which guide the video generation with a video diffusion transformer. The output is obtained from the video VAE decoder $\mathcal{D}_\mathrm{vid}$.
    For LiDAR generation, the sparse LiDAR UNet takes occupancy grids and sensor rig data as inputs, which are then passed to the LiDAR head $\mathcal{D}_\mathrm{lid}$ for multi-view LiDAR generation.\looseness=-1
    }
    \label{fig_overall}
    \vspace{-0pt}
\end{figure*}

\section{Methodology}

In this section, we introduce \ours, a unified framework designed 
for large-scale 4D dynamic scene generation of semantic occupancy, video, and LiDAR data. The framework upgrades UniScene on both the training data (\ie, Table~\ref{tab_dataset}) and model architecture (\ie, Figure~\ref{fig_overall}) to generate diverse large-scale 4D semantic occupancy generation, which is subsequently leveraged as conditional guidance for video and LiDAR generation.\looseness=-1

\noindent\textbf{Overview.}  
As illustrated in Figure~\ref{fig_overall}, we decompose the complex task of large-scale driving scene generation into an occupancy-centric hierarchical structure.  
Specifically, \ours\ first takes an optional bird's-eye-view (BEV) layout and noise volume as inputs to generate the expanded large-scale global semantic occupancy with a spatial occupancy Diffusion Transformer (DiT), which is further transformed into location-specific local temporal occupancy sequences using a temporal occupancy DiT (Section~\ref{sec_occ}). 
The resulting occupancy representation then acts as conditional guidance for subsequent video and LiDAR generation.  
For video generation, the occupancy is converted into 3D Gaussian primitives, which are rendered into 2D semantic and depth sparse point maps to guide the Video Diffusion Transformer (Section~\ref{sec_video}).  
For LiDAR generation, we propose a sensor-specific embedding scheme that integrates with the LiDAR Sparse UNet to learn occupancy priors with explicit sensor information for flexible and realistic LiDAR pattern simulation (Section~\ref{sec_lidar}).

\begin{figure}[!t]
    \centering
    \includegraphics[width=0.99\linewidth]{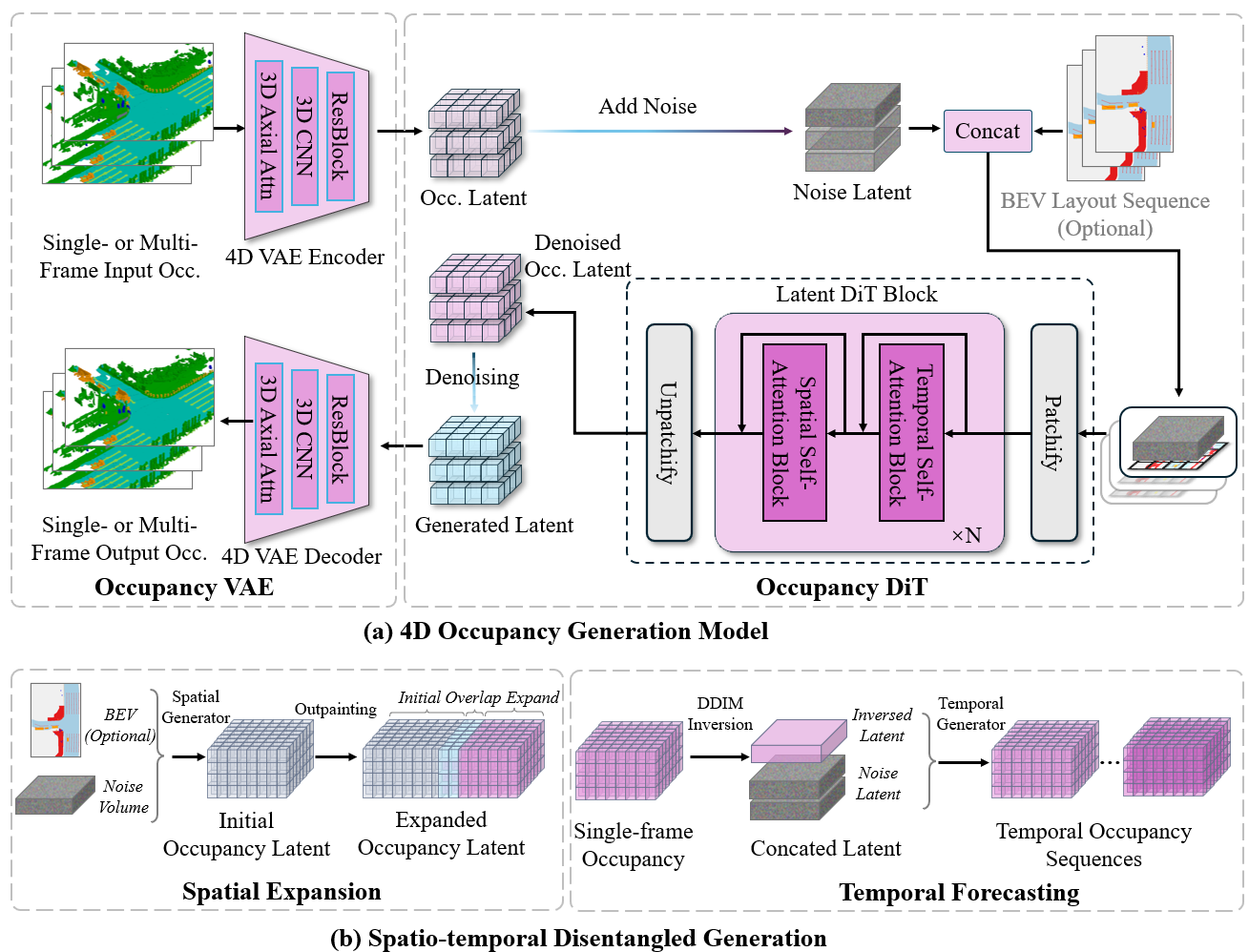}
    \caption{ (a) {Architecture of the occupancy generation model}, which integrates a 4D occupancy VAE and an occupancy Diffusion Transformer (DiT). 
     (b) {Spatio-temporal Disentangled Generation pipeline.}    
}
    \label{fig_occupancy}
\end{figure}

\subsection{4D Occupancy Generation}\label{sec_occ} 

The occupancy generation model mainly comprises a 4D occupancy VAE and an occupancy DiT. The occupancy generator supports both single-frame and multi-frame generation, with the option to incorporate a BEV layout as conditional guidance.
A Spatio-temporal disentangled modeling strategy is introduced to decouple 4D occupancy scene synthesis into two complementary tasks of spatial expansion and temporal forecasting.

\subsubsection{Occupancy VAE and DiT}
The architecture details of the 4D occupancy VAE and the occupancy DiT are shown in Figure~\ref{fig_occupancy} (a), which support controllable generation with BEV layouts or direct generation from pure noise.

\noindent\textbf{Occupancy VAE.} 
The occupancy VAE is designed to compress semantic occupancy data into a compact latent space, enhancing computational efficiency. 
Temporal information is incorporated during both the encoding and decoding processes to ensure consistent modeling. Specifically, different from the 2D-based processing in UniScene~\cite{li2024uniscene}, our occupancy VAE encoder is composed of a 3D CNN encoder enhanced and a 3D axial attention layer, which transforms a 3D semantic occupancy $\mathbf{O} \in \mathbb{R}^{H \times W \times D}$ within an occupancy sequence into a BEV representation $\mathbf{\hat{O}} \in \mathbb{R}^{H \times W \times DC'}$ by assigning each category a learnable class embedding $C'$. This occupancy VAE encoder extracts a continuous latent feature with a down-sampled resolution $\mathbf{Z}_\text{occ} \in \mathbb{R}^{C \times h \times w}$. Here, $h = \frac{H}{d}$ and $w = \frac{W}{d}$, where $d$ represents the down-sampling factor.
During decoding, the VAE reconstructs the latent feature sequence $\mathbf{z}_\text{occ}^\text{seq} \in \mathbb{R}^{T \times C \times h \times w}$. A 3D CNN network with a 3D axial attention layer is employed to up-sample the latent feature sequence into a BEV representation occupancy sequence $\mathbf{\hat{O}}^\text{seq} \in \mathbb{R}^{T \times H \times W \times DC'}$. This sequence is reshaped to $\mathbb{R}^{THW \times D \times C'}$ and processed through a dot product with the class embeddings to compute the logits scores. During training, the logits scores and one-hot labels are used to calculate the learning loss~\cite{berman2018lovasz}. In the inference phase, the final reconstructed occupancy sequence $\mathbf{O}^\text{seq} \in \mathbb{R}^{T \times H \times W \times D}$ is determined by applying the $\texttt{argmax}$ operation to the logits.

Following~\cite{zheng2023occworld}, we train the VAE using a combination of cross-entropy loss $\mathcal{L}_\text{CE}$, Lovász-softmax loss $\mathcal{L}_\text{LS}$, and Kullback–Leibler (KL) divergence loss $\mathcal{L}_\text{KL}$. The overall training objective is:
\begin{equation}
\begin{aligned}
   \mathcal{L}_\text{occ}^\text{vae} &= \mathcal{L}_\text{CE} + \lambda_1 \mathcal{L}_\text{LS} + \lambda_2 \mathcal{L}_\text{KL},
\end{aligned}
\end{equation}
where $\lambda_1$ and $\lambda_2$ denote the respective loss weights. 
Experimental validation is provided in Section~\ref{exp_occ}.

\noindent\textbf{Occupancy DiT.}  
The occupancy DiT is designed to denoise occupancy latent sequence features derived from noisy occupancy latents, optionally conditioned on BEV layout sequences. 
When BEV layout sequences are available, a unified patchify module is introduced to align the BEV layout with the occupancy latent features for fine-grained explicit control. Specifically, the BEV layout at time step $i$ is downsampled into $\mathbf{B}_{down}^{i} \in \mathbb{R}^{(C_b) \times h \times w}$ to match the spatial dimensions of the latent feature $\mathbf{Z}_\text{occ}^{i} \in \mathbb{R}^{(C_o) \times h \times w}$. These features are concatenated, resulting in $\mathbf{Z}_\text{cat} \in \mathbb{R}^{(C_o + C_b) \times h \times w}$. A unified patch embedder then transforms this concatenated latent into a sequence of unified latent tokens $\mathbf{Z} \in \mathbb{R}^{L \times E_d}$, where $L$ denotes the number of patches and $E_d$ represents the embedding dimension.\looseness=-1

The backbone of the occupancy DiT is the Spatial-Temporal Latent Diffusion Transformer, which consists of stacked spatial and temporal transformer blocks~\cite{ma2024latte}. The spatial blocks aggregate features across different positions within the same latent, while the temporal blocks capture dependencies across latent frames at the same spatial position. To encode relative spatial and temporal relationships, 2D positional embeddings and 1D temporal embeddings are incorporated. The output of the backbone, with dimensions $\mathbb{R}^{T \times L \times E_d}$, is passed through an unpatchify layer to produce a denoised occupancy latent sequence of size $\mathbb{R}^{T \times H \times W \times D}$.

During training, the BEV layout condition is randomly dropped with a probability of 0.1, enabling the diffusion model to learn unconditional generation. In the sampling phase, the classifier-free guidance scale is set to 1.0 by default when BEV layouts are available.
The training objective follows~\cite{peebles2023scalable}, minimizing the mean squared error between the predicted and target noise at each diffusion step:
\vspace{-0pt}
\begin{equation}
\mathcal{L}_\text{occ}^\text{dit} = \mathbb{E}\left[\sum_{i=1}^T\left \| \boldsymbol{f}_\text{dit}\left(\boldsymbol{z}_\text{occ}^{i}, \mathbf{B}^{i}\right)- \boldsymbol{\epsilon}_\text{n}^i \right\|^2 \right],
\end{equation}
where $\boldsymbol{f}_\text{dit}(\cdot)$ represents the model output, and $\boldsymbol{z}_\text{occ}^{i}$ denotes the noisy latent input at the $i^\text{th}$ frame.

\subsubsection{Spatio-temporal Disentangled Generation}

To address the complexity of generating dynamic large-scale 4D occupancy scenes, we decompose the task into two distinct components: spatial expansion and temporal forecasting.\looseness=-1

\noindent\textbf{Disentangled Data Construction.}  
To achieve this decomposition, the occupancy generation model is initially trained on the entire Nuplan-Occ dataset and subsequently fine-tuned using spatio-temporal disentangled data to separately obtain the spatial occupancy generator and the temporal occupancy generator. The spatio-temporal disentangled data is constructed according to the vehicle status. Specifically, the spatial data $\mathcal{S}_{\text{patial}}$ is constructed by filtering the Nuplan dataset to include only those scenes where the ego vehicle moves. Conversely, the temporal data $\mathcal{T}_{\text{emporal}}$ is constructed by filtering the dataset to include scenes where the ego vehicle is stationary while surrounding vehicles remain moving. 
The mathematical formulation of the data construction strategy is as follows:

\begin{equation}
\mathcal{S}_{\text{patial}} = \left\{ x \in \mathcal{D} \,\middle|\, v_{\text{ego}}(x) > \theta_e \right\}
\end{equation}

\begin{equation}
\mathcal{T}_{\text{emporal}} = \left\{ x \in \mathcal{D} \,\middle|\, v_{\text{ego}}(x) < \theta_e \land v_{\text{other}}(x) > \theta_o \right\}
\end{equation}

where $\mathcal{D}$ denotes the entire Nuplan dataset, and $x$ represents the filtered scenes. The function $v_{\text{ego}}(\cdot)$ extracts the speed of the ego vehicle from sensor data, while $v_{\text{other}}(\cdot)$ computes the speed of surrounding traffic vehicles from BEV layout sequences. The parameter $\theta_e$ is the speed threshold for the ego vehicle, distinguishing static from dynamic driving scenarios.
$\theta_o$ is the speed threshold for surrounding vehicles.
The operator $\land$ represents the logical conjunction, indicating that both conditions should be satisfied simultaneously.
Through this strategy, the spatial generator learns to capture dynamic scenes with consistent spatial relationships, while the temporal generator focuses on modeling vehicle motions with stable temporal dependencies.

\noindent\textbf{Spatial Expansion and Temporal Forecasting.} 
As shown in Figure~\ref{fig_occupancy} (b), the spatial expansion and temporal forecasting are achieved using occupancy generators that share the same architecture but are applied to different tasks.
For spatial expansion, the initial occupancy is either generated from a noise volume or guided by an input BEV layout. A 3D outpainting strategy is then employed to enable seamless scene expansion by conditioning on the initial occupancy. Specifically, to cover regions targeted for expansion, our diffusion model generates a novel 3D occupancy latent that partially overlaps with the original occupancy latent. A 3D occupancy latent mask is utilized to define the regions to be outpainted, while the known conditional occupancy latent is derived from the intersection of the original and expanded regions. This 3D outpainting process can be iteratively repeated to generate scenes of theoretically infinite size.\looseness=-1

For temporal occupancy sequence forecasting, the occupancy generation model is adapted into a temporal generative forecasting framework (temporal generator) that predicts $T_f$ future frames based on $T_c$ conditional frames, leveraging the spatial filter data for training. 
Specifically, during the training phase, the conditional occupancy latent (inversed latent) is obtained by encoding the selected single-frame occupancy with DDIM inversion, and concatenated with noise volumes without the BEV layouts. The unified latent representation for both $T_c$ (conditional) and $T_f$ (future) frames are then processed by the DiT backbone. The model outputs denoised occupancy latent frames for both $T_c$ and $T_f$, but the loss is computed exclusively on the $T_f$ frames.
As shown in Figure~\ref{fig_occupancy}, in the inference phase, the $T_f$ frames are initialized with pure noise, while the $T_c$ frame is initialized using the conditional occupancy latent sampled from the occupancy VAE. 
To align with previous studies~\cite{zheng2023occworld, wei2024occllama}, the default number of future occupancy frames $T_f$ is set to $6$. 
To enable long-term occupancy sequence generation, we leverage the roll-out strategy to facilitate multi-round generation~\cite{li2024uniscene,gao2025vista}.
To enhance computational efficiency and reduce reliance on conditional frames, we configure $T_c$ to $1$, instead of the $T_c = 5$ used in earlier works~\cite{zheng2023occworld, wei2024occllama}.

\begin{figure}[!t]
    \centering
    \includegraphics[width=0.99\linewidth]{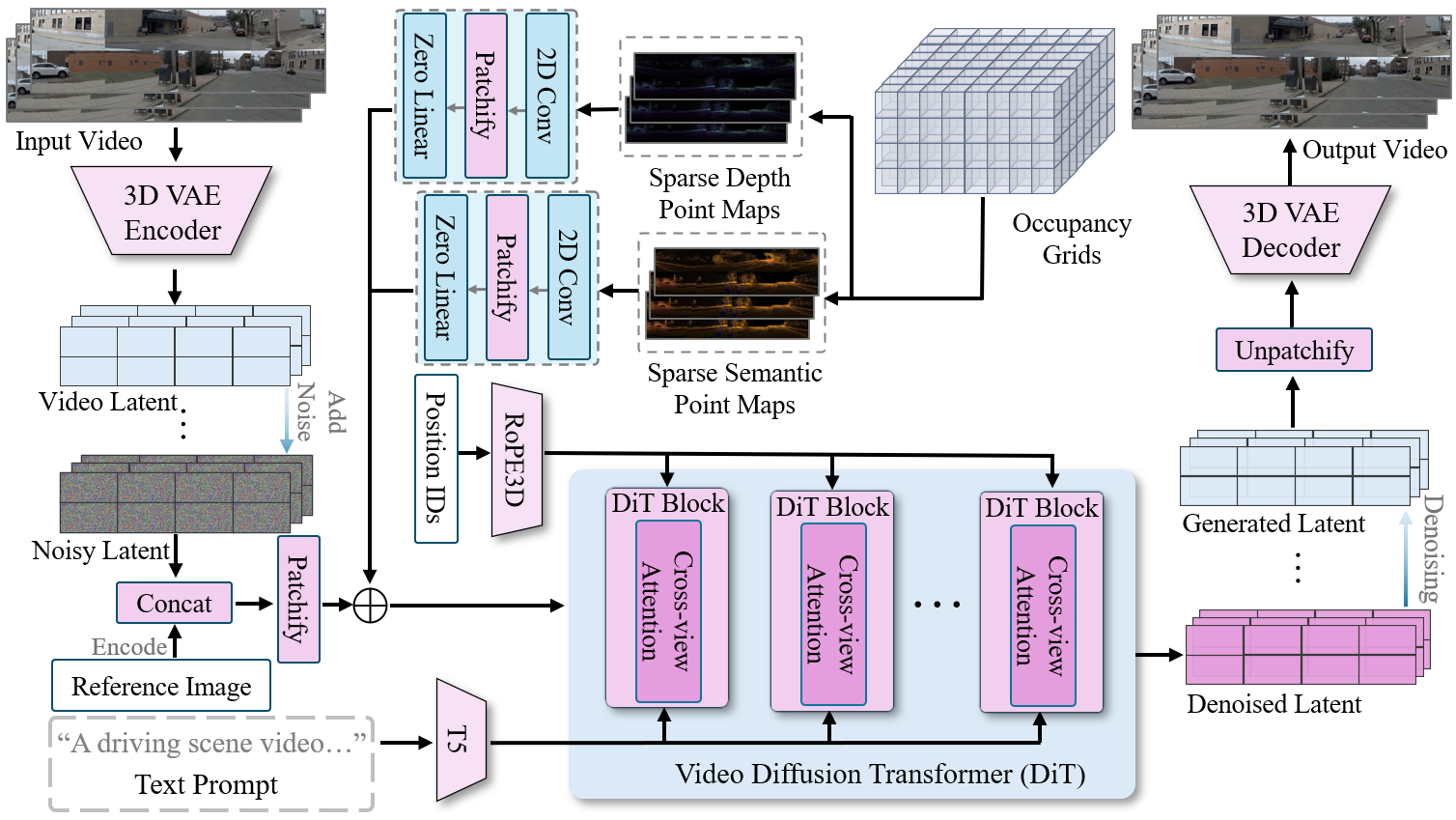}
    \caption{
    {The architecture of the video generation model}, which consists of a 3D video VAE and a video DiT. 
}
    \label{fig_video}
\end{figure}

\subsection{Video Generation with Gaussian Point Map}\label{sec_video}
As shown in Figure~\ref{fig_video}, the video generation model mainly consists of a 3D video VAE and a video diffusion Transformer (DiT), which synthesizes multi-view driving videos conditioned on occupancy-based rendering maps, reference images, and text prompts.\looseness=-1

\subsubsection{Video VAE and DiT}

To enable high-fidelity and scalable video generation, we employ a 3D causal VAE and a video DiT. Specifically, the 3D causal VAE is implemented following CogVideoX~\cite{yang2024cogvideox} and initialized with pre-trained weights. It provides an 8×8×4 compression ratio and outputs latent features with 16 channels.
Compared to the SVD~\cite{blattmann2023stable} VAE used in UniScene~\cite{li2024uniscene}, the 3D causal VAE offers greater efficiency through 4× temporal compression.
Furthermore, instead of the UNet architecture employed in UniScene, we adopt the 3D video DiT following Open-Sora Plan~\cite{lin2024open}, which enables better scalability and facilitates more effective training on the Nuplan \cite{Nuplan} dataset. 
Within the video DiT, 
The cross-view attention~\cite{gao2023magicdrive,wu2023tune} is added in each DiT block to facilitate multi-view consistency.
3D rotational position encoding (RoPE) is employed for capturing relative positional relationships rather than relying on absolute positions following~\cite{lin2024open,su2024roformer}.

\begin{figure}[!t]
    \centering
    \includegraphics[width=0.99\linewidth]{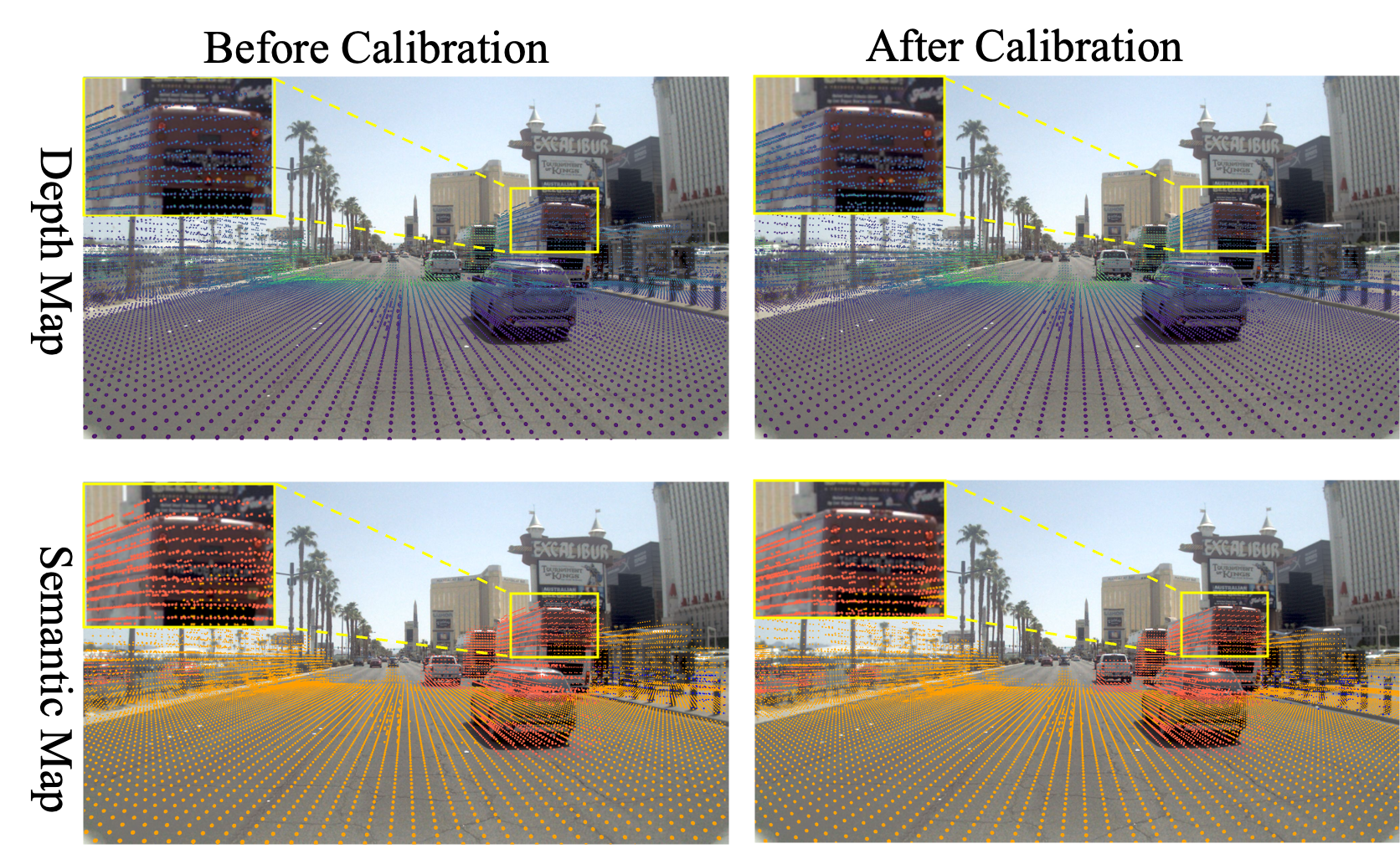}
    \caption{ 
    Robust calibration of Gaussian rendered point maps with unscented transform (UT). The calibration strategy effectively aligns the RGB image with the rendered semantic and depth maps, as highlighted by the buses and streetlight poles in the yellow bounding box.
    }
    \label{fig_gs_cal}
\end{figure}

\begin{figure}[!t]
    \centering
    \includegraphics[width=0.99\linewidth]{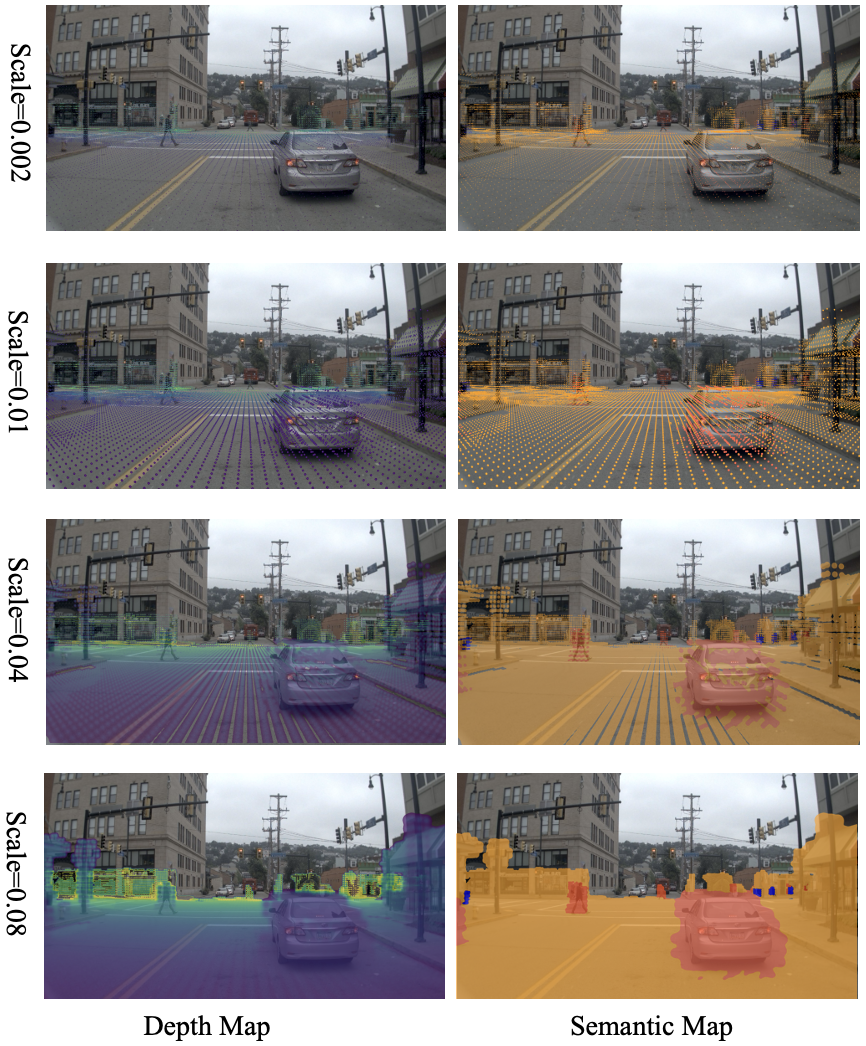}
    \caption{
      Gaussian-based sparse point map rendering with different scales. 
      The rendering maps with the default scale (0.01) provide sufficient and robust conditional priors.}
    \label{fig_gs_scale}
\end{figure}

\subsubsection{Gaussian Point Map Rendering}

While the Gaussian-based joint rendering strategy employed in UniScene improves performance by bridging the representational gap between occupancy grids and multi-view video, it does not account for sensor calibration misalignment and noise, which may degrade its effectiveness.\looseness=-1

\noindent\textbf{Sparse Gaussian Point Map Representation.} To address this limitation, we introduce a sparse Gaussian point map rendering strategy that provides robust semantic and geometric guidance, enabling high-quality and temporally consistent video generation. The experimental evaluation of this strategy is summarized in Table~\ref{table_ar_video}.

Specifically, input semantic occupancy grids are jointly rendered into multi-view semantic and depth sparse point maps using forward Gaussian splatting~\cite{kerbl20233d,zhou2024hugs}. Given an input semantic occupancy grid of shape $\mathbb{R}^{H \times W \times D}$, we first convert it into a set of 3D Gaussian primitives $\mathcal{G} = \{G_i\}_{i=1}^{N}$, where each $G_i$ corresponds to the center and semantic label of its respective voxel. 
Each Gaussian primitive encodes attributes including position $\mu$, semantic label $s$, opacity $\alpha$, and covariance $\Sigma$.
To ensure precise correspondence between the rendered sparse maps and the multi-view images, the scale of the Gaussian primitives is set to a relatively small value (default as 0.01). The impact of varying Gaussian primitive scales is illustrated in Table~\ref{table_ar_video}
Subsequently, the depth map $\mathbf{D}$ and semantic map $\mathbf{S}$ are rendered via tile-based rasterization~\cite{kerbl20233d}, analogous to color rendering:

\vspace{-0pt}
\begin{equation} 
\mathbf{D}=\sum_{i \in {N}} {d}_i \alpha_i^{\prime} \prod_{j=1}^{i-1}\left(1-\alpha_j^{\prime}\right),
\end{equation}
\vspace{-0pt}
\begin{equation} 
\mathbf{S}=\texttt{argmax} \left( \sum_{i \in {N}} \texttt{onehot}(s_i) \alpha_i^{\prime} \prod_{j=1}^{i-1}\left(1-\alpha_j^{\prime}\right) \right),
\end{equation}
where $d_i$ denotes the depth value, and $\alpha^{\prime}$ is derived from the projected 2D Gaussian and the 3D opacity $\alpha$.\looseness=-1

\noindent\textbf{Robust Calibration with Unscented Transform.} Moreover, to address sensor calibration misalignment and noise in Gaussian-based joint rendering, we introduce a robust unscented transform (UT) integrated rendering pipeline. While forward Gaussian splatting efficiently renders depth and semantic maps from occupancy grids, traditional Elliptical Weighted Average (EWA) splatting relies on linearized projections that degrade under significant camera distortions. To ensure precise alignment with multi-view imagery—particularly for datasets like Nuplan~\cite{Nuplan} with pronounced lens distortion—we integrate the Unscented Transform~\cite{wu20253dgut} into our projection pipeline.
 
Given a 3D Gaussian primitive $G_i$ with position $\bm{\mu} \in \mathbb{R}^3$ and covariance $\bm{\Sigma} \in \mathbb{R}^{3 \times 3}$, UT approximates its distribution using $2N+1=7$ sigma points ($N=3$ dimensions). These points $\mathcal{X} = \{\bm{x}_k\}_{k=0}^{6}$ are computed as:  
\begin{equation} 
\bm{x}_k = 
\begin{cases} 
\bm{\mu} & k=0 \\
\bm{\mu} + \sqrt{(3 + \lambda)} \cdot \bm{L}_{[:,k]} & k=1,2,3 \\
\bm{\mu} - \sqrt{(3 + \lambda)} \cdot \bm{L}_{[:,k-3]} & k=4,5,6 
\end{cases}
\end{equation} 
where $\bm{L}$ is the Cholesky factor of $\bm{\Sigma}$ (\ie, $\bm{\Sigma} = \bm{L}\bm{L}^\top$), and $\lambda = \alpha^2(3 + \kappa) - 3$. Hyperparameters $\alpha=1.0$, $\beta=2.0$, and $\kappa=0.0$ control point spread and distribution prior knowledge following~\cite{wu20253dgut}. 
Each sigma point is projected onto the image plane via the nonlinear camera model $\bm{v}_k = g(\bm{x}_k)$, which natively incorporates radial/tangential distortion and rolling shutter effects. The mean $\bm{v}_\mu$ and covariance $\bm{\Sigma}'$ of the projected 2D conic are then estimated:  
\begin{equation} 
\bm{v}_{\mu} = \sum_{k=0}^{6} w_k^\mu \bm{v}_k, \quad 
\bm{\Sigma}' = \sum_{k=0}^{6} w_k^\Sigma (\bm{v}_k - \bm{v}_{\mu})(\bm{v}_k - \bm{v}_{\mu})^\top
\end{equation} 
with weights $w_k^\mu$ and $w_k^\Sigma$ defined as:  
\begin{equation} 
\begin{aligned}
w_0^\mu &= \lambda / (\lambda + 3),  & w_{1:6}^\mu &= 1 / \big(2(\lambda + 3)\big) \\
w_0^\Sigma &= w_0^\mu + (1 - \alpha^2 + \beta), & w_{1:6}^\Sigma &= w_{1:6}^\mu.
\end{aligned}
\end{equation}

{As shown in Figure~\ref{fig_gs_cal}, the UT-integrated robust rendering pipeline seamlessly bridges the gap between occupancy grids and multi-view video under challenging sensor conditions, enabling accurate semantic-geometric alignment.
Additionally, the visualization results of the rendered semantic and depth maps with different Gaussian scales are illustrated in Figure~\ref{fig_gs_scale}.
Compared to dense rendering with a larger Gaussian scale, the sparse point maps generated with a smaller Gaussian scale exhibit more robust and precise alignment with the multi-view images.}
These rendering maps are subsequently processed through 2D convolutions, followed by spatial and temporal downsampling. They are then patched and aligned with the latent feature space. A linear layer initialized with zeros is applied before fusing these features with the latent features. This design preserves the pre-trained capabilities of the video diffusion transformer while maintaining its generative potential.

\subsubsection{Video Training Loss} 

The video training loss is defined following established approaches~\cite{blattmann2023stable,gao2024vista}, formulated as:

\vspace{-0pt}
\begin{equation}
\mathcal{L}_\text{vid} = \mathbb{E} \left[ \sum_{i=1}^T \left\| \boldsymbol{f}_\text{vid} \left( \boldsymbol{z}_\text{vid}^i, t, \boldsymbol{z}_{c}, \mathbf{D}^{i}, \mathbf{S}^{i} \right) - \boldsymbol{z}_0^i \right\|^2 \right],
\end{equation}
where 
$\boldsymbol{f}_\text{vid}(\boldsymbol{z}_\text{vid}^i, t, \boldsymbol{z}_{c}, \mathbf{D}^{i}, \mathbf{S}^{i})$ denotes the output of the video generation model.  
$\mathbf{D}^{i}$ and $\mathbf{S}^{i}$ represent the depth and semantic maps of the $i^\text{th}$ video frame, respectively.  
$t$ denotes the input text prompt.  
$\boldsymbol{z}_0^i$ and $\boldsymbol{z}_\text{vid}^i$ denote the ground truth and noisy input latent representations at frame $i$, respectively.  
$\boldsymbol{z}_c$ represents the conditional reference frame.

\subsection{LiDAR Generation with View Decoupling}\label{sec_lidar} 

\begin{figure}[!t]
    \centering
    \includegraphics[width=0.99\linewidth]{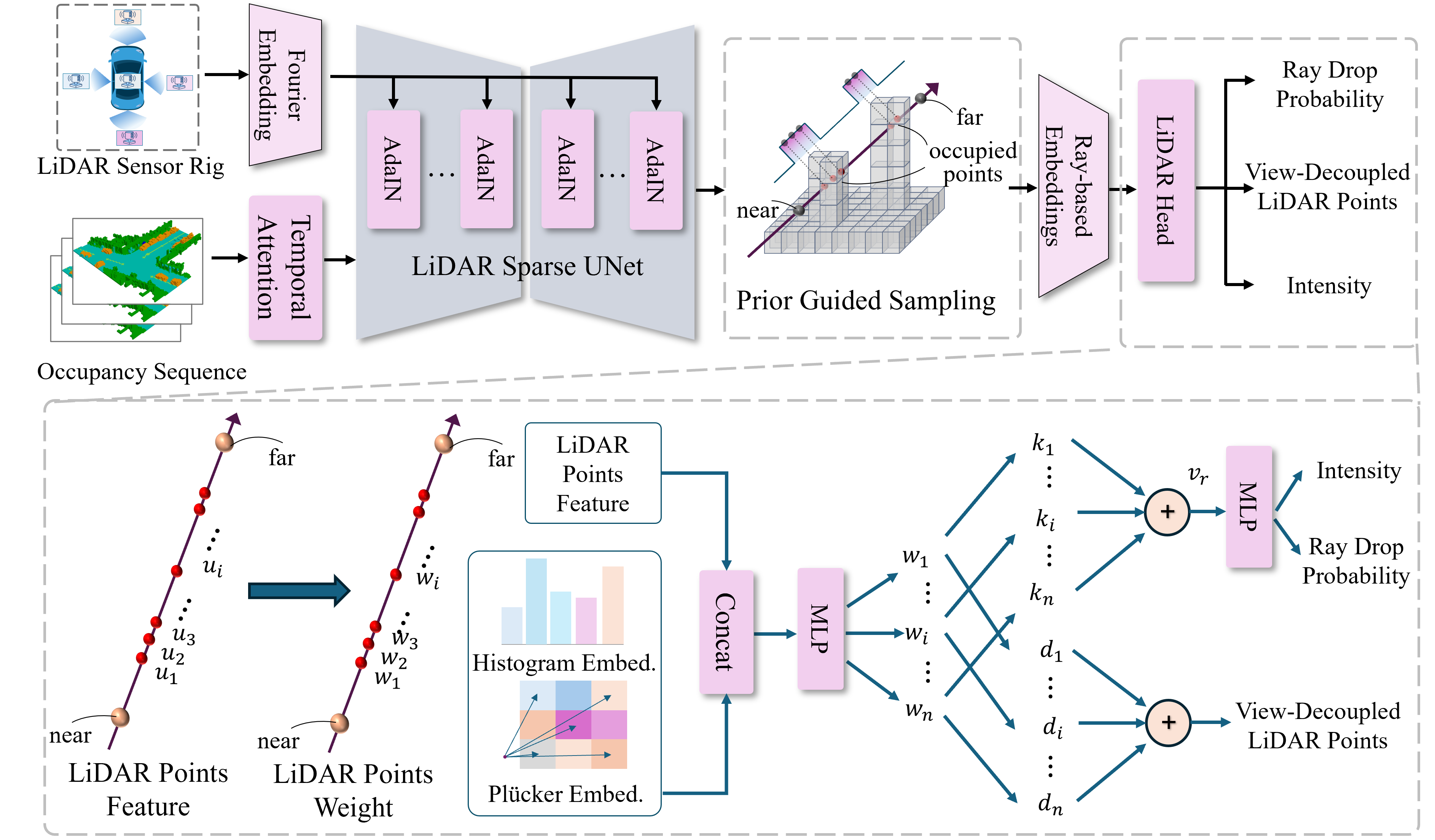}
    \caption{{The architecture of the LiDAR generation model}, which takes occupancy sequences and LiDAR sensor rig as inputs, and produces view-decoupled LiDAR points. 
    }
    \label{fig_lidar}
\end{figure}

As illustrated in Fig.~\ref{fig_lidar}, the LiDAR generation process begins by encoding the input occupancy into sparse voxel features using a Sparse UNet~\cite{shi2020points}. These features are then utilized to generate LiDAR points through a sparse sampling process guided by occupancy priors.
To precisely and flexibly simulate the LiDAR patterns, a sensor-specific embedding scheme is proposed to explicitly leverage LiDAR sensor rig data. Moreover, a smoothness loss term is introduced to facilitate the continuity of simulated LiDAR scanlines and reduce the noise of discrete LiDAR points.

\subsubsection{Occupancy Guided Sparse Modeling}
To address the inherent sparsity and detailed geometry of semantic occupancy, we introduce a prior-guided sparse modeling approach that enhances computational efficiency by avoiding unnecessary computations on unoccupied voxels. The input semantic occupancy grids are first processed with a Sparse UNet~\cite{shi2020points} to aggregate contextual features. Subsequently, uniform sampling is performed along LiDAR rays, denoted as $\mathbf{r}$, to generate a sequence of points represented as $s$.
As shown in Figure~\ref{fig_lidar}, prior-guided sparse sampling is facilitated by assigning a probability of 1 to points within occupied voxels and 0 to all other points, thereby defining a probability distribution function (PDF). Based on this PDF, $n$ points $\{\mathbf{r}_i = {o} + s_i{v} \ (i=1,...,n)\}$ are resampled, where ${o}$ represents the ray origin and ${v}$ denotes the normalized ray direction. Then, geometric features $\mathbf{e}_g$ of each sampled point can be extracted from the sparse tensor $\mathcal{X}_{occ}$ output by the Sparse UNet using bilinear interpolation:
\begin{equation}
    \mathbf{e}_g = \text{Interp}(\mathbf{r}, \mathcal{X}_{occ}).
\end{equation}
Moreover, two additional ray feature embeddings are incorporated to facilitate high-quality simulation.

\noindent\textbf{Histogram Embedding for Ray Features.}
To fully utilize the occupancy-based prior, we compute a per-ray histogram feature encoding the occupancy distribution of sampled points along each ray. Specifically, we partition the ray uniformly into 64 bins and assign each sampled point to its corresponding bin. The bin counts are accumulated and normalized, yielding a 64-dimensional histogram vector $\mathbf{h} \in \mathbb{R}^{64}$.
To reduce the dimensionality of the histogram while preserving its information, we introduce 64 learnable embeddings $\mathbf{E}_h \in \mathbb{R}^{64 \times 16}$, each of dimensionality 16. The final histogram feature is computed as:
\begin{equation}
\mathbf{e}_h = \mathbf{E}_h^T \mathbf{h}.
\end{equation}

\noindent\textbf{Plücker Embedding for Ray Features.}
Features sampled directly from sparse tensors primarily capture local geometric information from the voxel containing each sampling point. To incorporate ray-specific information and enhance feature consistency across neighboring rays, we augment the original features with Plücker coordinates. Specifically, for a ray $\mathbf{r} = \mathbf{o} + t\mathbf{d}$, its Plücker embedding is defined as: 
\begin{equation}
    \mathbf{e}_p = \text{Cat}(\mathbf{d}, \mathbf{o} \times \mathbf{d}),
\end{equation}
which jointly encodes the ray's origin and direction. 

Finally, the feature for each sampled point is obtained by concatenating the geometry feature, Plücker embedding, and histogram embedding, denoted as:
\begin{equation}
    \mathbf{f} = \text{Cat}(\mathbf{e}_g, \mathbf{e}_h, \mathbf{e}_p).
\end{equation}

\noindent\textbf{LiDAR Generation Head.}
Building on ray-based volume rendering techniques from prior works~\cite{unisim, Barron_2023_ICCV, wang2021neus}, the features of each resampled point are processed through a multi-layer perceptron (MLP) to predict the signed distance function (SDF) $f(s)$ and compute the associated weights $w(s)$.  
These predictions and weights are then used to estimate the depth of the ray via volume rendering:
\begin{equation}
    \beta_i = \max\left(\frac{\Phi_s(f(\mathbf{r}(s_i))) - \Phi_s(f(\mathbf{r}(s_{i+1})))}{\Phi_s(f(\mathbf{r}(s_i)))}, 0\right),
\end{equation}
\begin{equation}
    w(s_i) = \prod_{j=1}^{i-1}(1-\beta_j)\beta_i, \quad h = \sum_{i=1}^n w(s_i)s_i,
\end{equation}
where $\Phi_s(x) = (1 + e^{-sx})^{-1}$ and $h$ represents the rendered depth value.
The ray feature, ${v}_r$, is obtained by performing a weighted summation of the features of all points along the ray, expressed as:
\begin{equation}
    v_r = \sum_{i=1}^{n} k_i = \sum_{i=1}^{n} w_i \cdot u_i.
\end{equation}
Finally, ${v}_r$ is processed through another MLP layer to simultaneously predict the intensity and drop probability of the LiDAR ray.\looseness=-1

To better simulate realistic LiDAR imaging, we incorporate two components: a reflection intensity head and a ray-dropping head. The reflection intensity head predicts the reflection intensity of the LiDAR beam for each ray. This is computed as a weighted sum of point features along the ray using weights $w(s)$, followed by an MLP for intensity estimation. The ray-dropping head estimates the probability that a ray is dropped due to undetected reflections.

\begin{figure}[!t]
    \centering
   \includegraphics[width=0.99\linewidth]{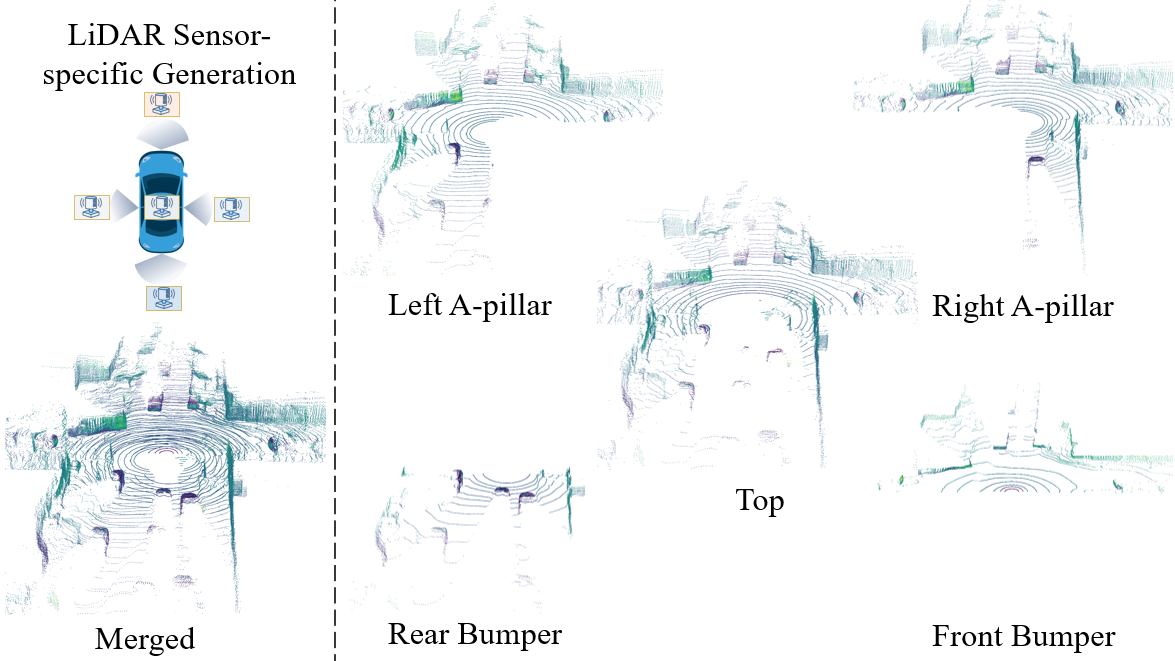}
        \caption{Sensor-specific Embedding for decoupled LiDAR generation, which estimates the extrinsic parameters of each LiDAR sensor to enable flexible pattern simulation. }
        \label{fig:lidar_decouple}
  
\end{figure}

\begin{figure}[!t]
    \centering
     \includegraphics[width=0.99\linewidth]{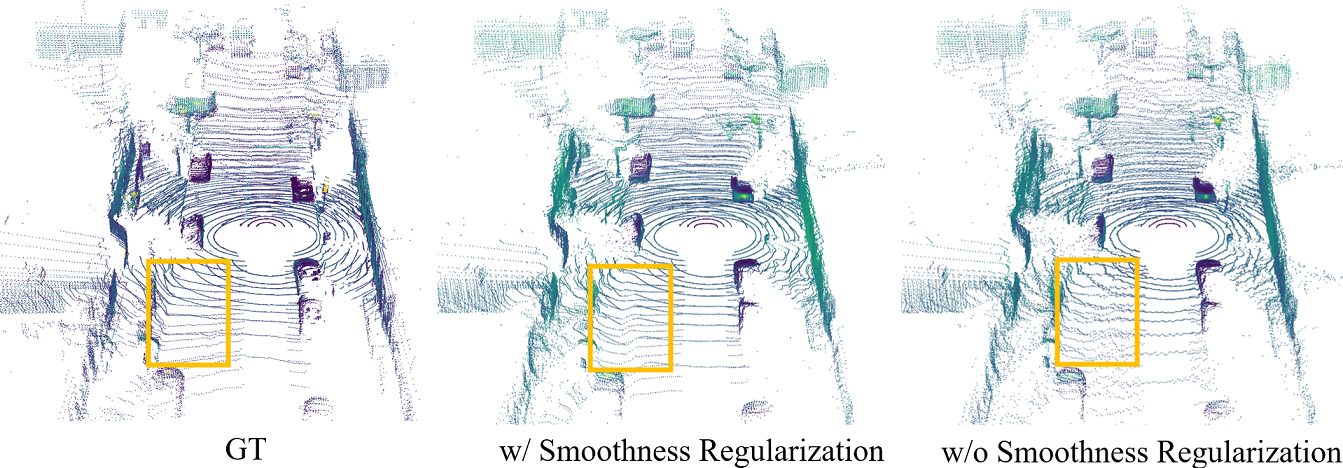}
        \caption{Visualization of the LiDAR ray smoothness regularization strategy. The explicit regularization strategy effectively produces more continuous and accurate simulation patterns, as highlighted in the road surface regions.}
        \label{fig:lidar_smooth}
\end{figure}

\subsubsection{Sensor-specific Embedding}

As illustrated in Figure~\ref{fig:lidar_decouple}, the Nuplan dataset provides five LiDAR sensors, which are originally merged together.
Straightforwardly modeling the LiDAR points of multiple sensors is non-trivial due to the mixed scanning patterns and the lack of extrinsic calibration parameters. 
To address this limitation and facilitate flexible simulation, we estimate the extrinsic parameters of each LiDAR sensor by leveraging the regular scanline pattern characteristic of LiDARs. However, these estimates still contain residual errors. Directly supervising the model using point clouds from all LiDARs may lead to conflicting gradients due to calibration inaccuracies.\looseness=-1

To make the network aware of the LiDAR rig configuration and enable flexible simulation of various LiDAR setups, we propose a Sensor-specific Embedding module. Specifically, we first apply Fourier encoding to the origin of each LiDAR sensor to obtain its corresponding LiDAR embedding $\mathbf{e}_l$. During training, we randomly select $n_l$ LiDARs, setting the embeddings of unselected LiDARs to zero. All LiDAR embeddings are then processed through a resampling network to generate a unified LiDAR rig embedding $\mathbf{f}_r$, which encapsulates the configuration of the entire LiDAR suite. Next, we inject $\mathbf{f}r$ into each Sparse U-Net block using AdaLN-Zero conditioning. Specifically, the conditioned output is computed as:
\begin{equation}
\mathcal{X}_{cond} = \mathcal{X} + \text{AdaLN}(\text{Conv3D}(\mathcal{X})).
\end{equation}

\subsubsection{Ray Smoothness Regularization}
We observe that due to the continuity of LiDAR scanlines, depth measurements in flat regions (\eg, roads and walls) exhibit smoothly varying patterns. While our Plücker embedding injection module ensures continuous and consistent features across neighboring scanlines, it lacks explicit regularization during training. Inspired by the depth smoothness regularization~\cite{godard2019digging,fu2018deep}, we propose a Ray Smoothness Regularization strategy.

Firstly, the estimated LiDAR point cloud is projected onto a range map $d \in \mathbb{R}^{H_d \times W_d}$, where $H_d$ corresponds to the elevation (pitch) dimension and $W_d$ to the azimuth dimension, with $d(i, j)$ denoting the depth of the LiDAR point at the corresponding pixel. In addition, for each ray, we compute a histogram based on the distribution of sampled points along the ray,  serving as a ray-specific feature. This histogram is also projected onto the range map, resulting in $h \in \mathbb{R}^{C_h \times H_d \times W_d}$, where $C_h$ is the number of histogram bins. We posit that rays with similar histograms should produce similar depth values, leading to the smoothness  regularization:
\begin{equation}
    \mathcal{L}_s=|\partial_xd|e^{-\partial_x h}.
\end{equation}
This encourages depth smoothness between rays with similar histogram features, while allowing for depth discontinuities at locations where histogram features change abruptly.
As illustrated in Figure~\ref{fig:lidar_smooth}, the explicit LiDAR ray smoothness regularization strategy effectively yields more continuous and accurate simulation patterns, as highlighted by the yellow bounding box in the road surface regions.

The overall training loss for LiDAR generation comprises four components: depth loss $\mathcal{L}_\text{depth}$, intensity loss $\mathcal{L}_\text{inten}$, ray-dropping loss $\mathcal{L}_\text{drop}$, and smoothL1 loss $\mathcal{L}_\text{smooth}$:
\begin{equation}
\begin{aligned}
\mathcal{L}_\text{lid} &= \mathcal{L}_\text{depth} + \lambda_1\mathcal{L}_\text{inten} + \lambda_2\mathcal{L}_\text{drop} + \lambda_3\mathcal{L}_\text{smooth},
\end{aligned}
\end{equation}
where $\lambda_1$, $\lambda_2$ and $\lambda_3$ are balancing coefficients.

\section{Experiments}

Our framework undergoes a two-stage training process implemented with PyTorch on 64 NVIDIA A100 GPUs. Initially, the occupancy generative models are trained using ground-truth labels. 
Subsequently, the occupancy generative model is fixed to generate occupancy grids from the BEV maps, while the video and LiDAR generation models are jointly trained with occupancy-based conditions. More details are provided in the supplementary materials.

\subsection{Main Results}

\begin{figure*}[!t]
    \centering
    \vspace{-0pt}
    \includegraphics[width=0.9\linewidth]{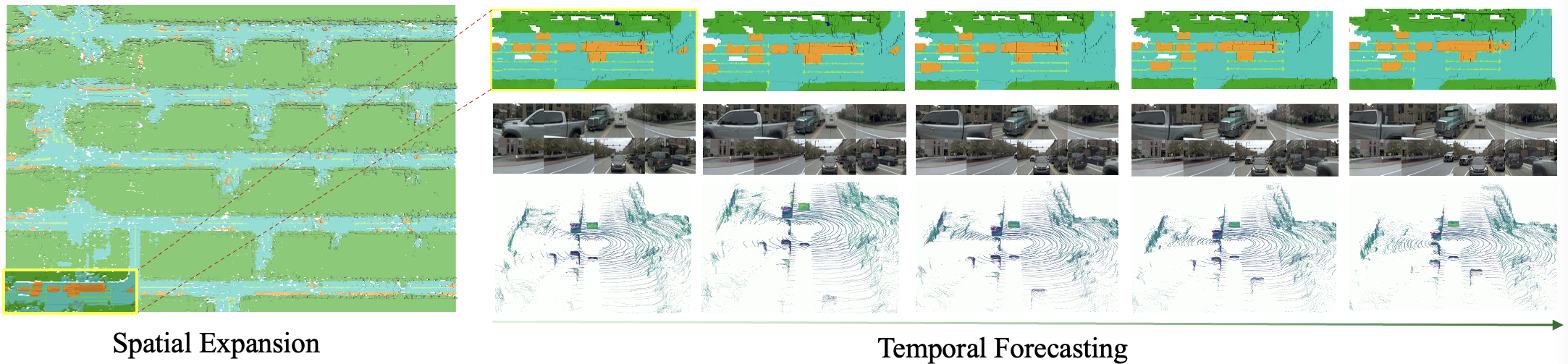}
    \vspace{-0pt}
    \caption{Visualization of scene expansion and forecasting results.
    UniScenev2 enables spatio-temporally disentangled generation, supporting both large-scale spatial expansion and future occupancy sequence prediction, while jointly producing multi-view video and LiDAR data in a unified pipeline.
    }
    \label{fig_teaser2}
    \vspace{-0pt}
\end{figure*}

\noindent\textbf{Scene Expansion and Forecasting.} 
The visualization results for scene expansion and forecasting are presented in Figure~\ref{fig_teaser2}.
UniScene v2 facilitates spatio-temporally disentangled generation, enabling large-scale driving scene expansion and multi-frame forecasting.
Moreover, the framework supports unified synthesis of corresponding 3D semantic occupancy, multi-view video streams, and LiDAR point clouds, demonstrating its capability for holistic 4D dynamic scene simulation.

\begin{table}[!t]
\vspace{-0pt}
\centering
\scriptsize
\renewcommand\tabcolsep{5.0pt}
\resizebox{0.99\linewidth}{!}{
\begin{tabular}{c|l|c|c|c}
\toprule
{Dataset} & {Method} & 
\begin{tabular}[c]{@{}c@{}}{Compression}\\{Ratio}\end{tabular} $\uparrow$ & 
{mIoU} $\uparrow$ & {IoU} $\uparrow$ \\ 
\midrule

\multirow{8}{*}{Mini}
& OccWorld ~\cite{zheng2023occworld} & 16 & 60.2 & 52.7  \\
& OccSora ~\cite{wang2024occsora} & {512} & 44.9 & 29.6 \\

& UniScene~\cite{li2024uniscene} & {32}  & {91.4}  & {84.0} \\
& UniScene~\cite{li2024uniscene} & {512}  & 61.3 & 59.2   \\ 
& UniScenev2 (Ours) & {32}  & \textbf{94.7} & \textbf{93.4}  \\
& UniScenev2 (Ours) & {512} & 62.4  & 69.8 \\   
\midrule

\multirow{2}{*}{Full} 
& UniScenev2 (Ours) & {32}  &  \textbf{98.5} & \textbf{97.8} \\
& UniScenev2 (Ours) & {512} &   70.8  & 70.5 \\    
\bottomrule
\end{tabular}
}
\vspace{-0pt}
\caption{
Quantitative evaluation for occupancy reconstruction on the Nuplan-Occ mini/full validation set.
The compression ratio is calculated following the methodology outlined in OccWorld~\cite{zheng2023occworld}.  
The baseline methods are evaluated on the mini validation set of Nuplan-Occ.
We additionally evaluate our method
on the full validation set.
}
\vspace{-0pt}
\label{tab_occ_rec}
\end{table}

\begin{table}[!t]
\vspace{-0pt}
\begin{center}
\scriptsize
\vspace{-0pt}
\renewcommand\tabcolsep{6.5pt}
\centering
\resizebox{0.99\linewidth}{!}{
\begin{tabular}{c|l|c|c|c}
\toprule
{Dataset} & {Method} & {mIoU $\uparrow$} & {F3D $\downarrow$} & {MMD $\downarrow$} \\
\midrule

\multirow{5}{*}{Mini}
& OccWorld~\cite{zheng2023occworld} & 17.52 & 164.23  & 12.56 \\
& OccSora~\cite{wang2024occsora} & 15.11 & 207.70 & 11.23\\
& UniScene~\cite{li2024uniscene}  & 22.64 & 130.72 & 9.60 \\  
&  UniScenev2 (Ours)  & \textbf{32.22} & \textbf{48.24} & \textbf{0.784} \\
\midrule
\multirow{1}{*}{Full} & UniScenev2 (Ours)  & \textbf{33.41} & \textbf{46.42} & \textbf{0.672} \\

\bottomrule
\end{tabular}
}
\vspace{-0pt}
\caption{{Quantitative evaluation for occupancy generation on the Nuplan-Occ mini/full validation set. The VAE compression ratio of 512 is utilized as the default setting.} 
}
\vspace{-0pt}
\label{occ_gen_fidelity}
\end{center}
\end{table}

\noindent\textbf{Occupancy Reconstruction and Generation.}\label{exp_occ}
As shown in Table~\ref{tab_occ_rec} and Table~\ref{occ_gen_fidelity}, the comparison results of occupancy evaluation are on the Nuplan-Occ mini validation set.
Moreover, we also provide the evaluation results of our method on the Nuplan-Occ full validation set.
As shown in Table~\ref{tab_occ_rec}, compared to the discrete compression with VQVAE in previous works~\cite{zheng2023occworld,wang2024occsora}, our continuous compression with VAE achieves remarkable reconstruction performance even under the high compression ratio of 512, 
surpassing OccWorld~\cite{zheng2023occworld} by 34.43\% in mIoU. 
Compared to UniScene~\cite{li2024uniscene}, our method improves 3.30 mIoU, which can be attributed to the 4D occupancy VAE that fully aggregates spatial and temporal context. 
The quantitative evaluation for occupancy generation is shown in Table~\ref{occ_gen_fidelity}.
Our method generates high-quality results with a default VAE compression ratio of 512, improving 14.70 mIoU and 9.58 mIoU compared to OccWorld~\cite{zheng2023occworld} and UniScene~\cite{li2024uniscene}, respectively.
Our method yields more complete and precise results compared to previous works.

\begin{table}[!t]
\vspace{-0pt}
\begin{center}
\scriptsize
\vspace{-0pt}
\renewcommand\tabcolsep{3.5pt}
\centering
\resizebox{0.99\linewidth}{!}{
\begin{tabular}{c|l|cc|cc}
\toprule
{Dataset} & {Method} & {Video} & {Multi-View} & {FID $\downarrow$} & {FVD $\downarrow$} \\
\midrule

\multirow{9}{*}{Mini}
& BEVGen~\cite{swerdlow2024street} & \textcolor{red}{\usym{2717}} & \textcolor{red}{\usym{2717}} & 29.84 & - \\
& DriveDreamer~\cite{wang2023drivedreamer} & \textcolor{ForestGreen}{\usym{2713}} & \textcolor{red}{\usym{2717}} & 21.94 & 427.65 \\
& MagicDrive~\cite{gao2023magicdrive} & \textcolor{ForestGreen}{\usym{2713}} & \textcolor{ForestGreen}{\usym{2713}} & 18.52 & 241.76 \\  
& Vista~\cite{gao2024vista} & \textcolor{ForestGreen}{\usym{2713}} & \textcolor{red}{\usym{2717}} & 11.64 & 108.50 \\ 
& Vista$^*$~\cite{gao2024vista} & \textcolor{ForestGreen}{\usym{2713}} & \textcolor{ForestGreen}{\usym{2713}} & 15.71  & 133.84   \\ 
& UniScene  & \textcolor{ForestGreen}{\usym{2713}} & \textcolor{ForestGreen}{\usym{2713}} & 9.84 & 89.36  \\
& UniScenev2 (Ours)   & \textcolor{ForestGreen}{\usym{2713}} & \textcolor{ForestGreen}{\usym{2713}} &  \textbf{8.32}  & \textbf{63.29} \\
\midrule

\multirow{1}{*}{Full}
& UniScenev2 (Ours) & \textcolor{ForestGreen}{\usym{2713}} & \textcolor{ForestGreen}{\usym{2713}} & \textbf{7.59} & \textbf{61.42}  \\
\bottomrule
\end{tabular}
}
\vspace{-0pt}
\caption{
Quantitative evaluation for video generation on the Nuplan-Occ mini/full validation set. We implement the multi-view variant of Vista$^*$~\cite{gao2024vista} with spatial-temporal attention~\cite{wu2023tune}.
}
\vspace{-0pt}
\label{tab_video}
\end{center}
\end{table}

\begin{figure}[!t]
    \centering
    \vspace{-0pt}
    \includegraphics[width=1.0\linewidth]{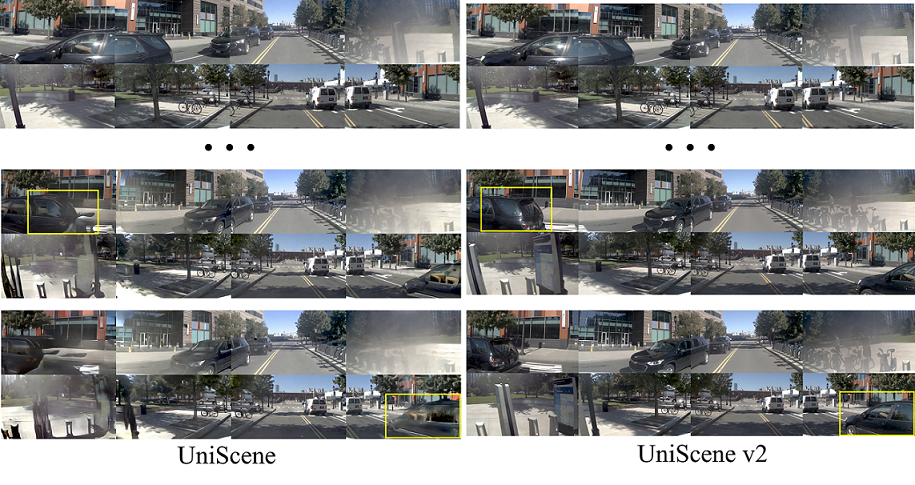}
    \vspace{-0pt}
    \caption{{Qualitative evaluation} for video generation. Our method produces more consistent and high-fidelity object structures.\looseness=-1}
    \label{video_compare}
    \vspace{-0pt}
\end{figure}

\noindent\textbf{Video Generation Results.}
The quantitative comparison of video generation is illustrated in Tab.~\ref{tab_video}. 
Our method supports multi-view video generation and outperforms all the other methods, achieving 8.32 FID and 63.29 FVD with ground truth occupancy, respectively.
As shown in Fig.~\ref{video_compare}, we compare our video generation results with UniScene~\cite{li2024uniscene}. 
Our approach demonstrates an obvious improvement in video generation quality, particularly in the structure quality of the moving vehicles. The notable enhancement is attributed to the robust conditional guidance derived from occupancy-based sparse point maps.\looseness=-1

\begin{table}[!t]
\vspace{-0pt}
\begin{center}
\scriptsize
\vspace{-0pt}
\renewcommand\tabcolsep{8.0pt}
\centering
\resizebox{1.00\linewidth}{!}{
\begin{tabular}{c|l|cc}
\toprule
{Dataset} & {Method} & {MMD} ($10^{-5}$)$\downarrow$ & {JSD} $\downarrow$ \\
\midrule

\multirow{5}{*}{Mini}
& Open3D~\cite{zhou2018open3d} & 15.429 & 0.116  \\
& LiDAR-Diffusion~\cite{ran2024towards} &19.940 & 0.161  \\
& UniScene~\cite{li2024uniscene} & 0.999 & 0.033  \\
& UniScenev2 (Ours)  & \textbf{0.457} & \textbf{0.028} \\

\midrule

\multirow{1}{*}{Full}
& UniScenev2 (Ours) & \textbf{0.575} & \textbf{0.032} \\
\bottomrule
\end{tabular}
}
\vspace{-0pt}
\caption{
{Quantitative evaluation for LiDAR Generation} on the Nuplan mini/full validation set.
}
\vspace{-0pt}
\label{tab_lidar}
\end{center}
\end{table}

\begin{figure}[!t]
    \centering
    \includegraphics[width=1.0\linewidth]{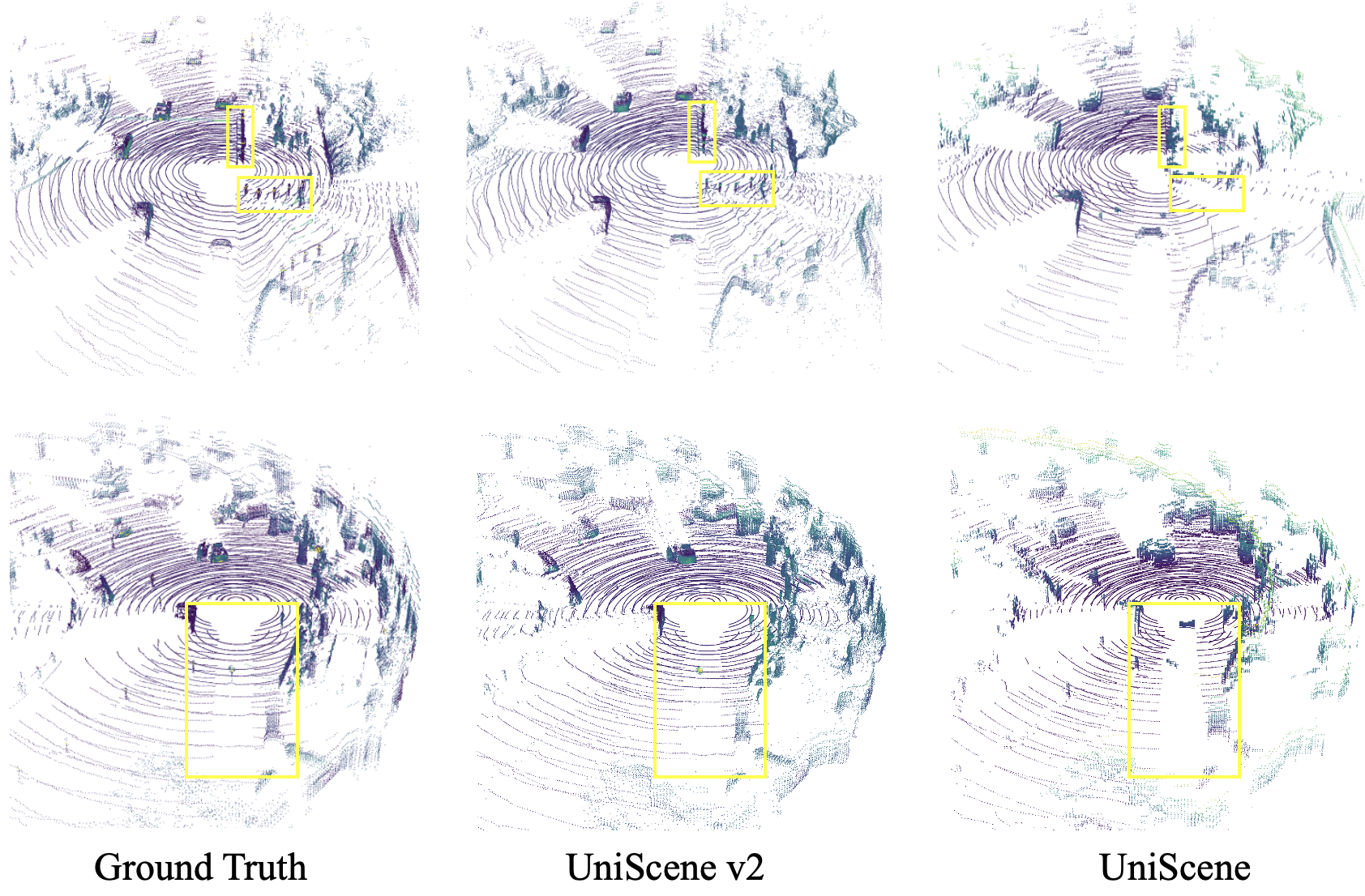}
    \vspace{-0pt}
    \caption{{Qualitative evaluation} for LiDAR generation. Our method generates precise scene layouts and structural details.}
    \label{lidar_compare}
    \vspace{-0pt}
\end{figure}

\noindent\textbf{LiDAR Generation Results.}
We compare our LiDAR generation model against Open3D~\cite{zhou2018open3d}, LiDAR-Diffusion~\cite{ran2024towards}, and UniScene~\cite{li2024uniscene} on the Nuplan mini/full validation set. For Open3D, we employ the library's ray-casting function to convert ground truth occupancy grids into corresponding LiDAR point clouds. LiDAR-Diffusion is implemented using its official repository and trained under the same conditions as our model. 
As presented in Tab.~\ref{tab_lidar}, our method achieves superior generation performance, surpassing UniScene by 54.25\% in MMD. Qualitative results are provided in Fig.~\ref{lidar_compare}. Compared to UniScene, our approach demonstrates a significant advantage in generating precise scene layouts and clear structural details.

\begin{table}[!t]
\begin{center}
\scriptsize
\vspace{-0pt}
\renewcommand\tabcolsep{12.0pt}
	\centering
   \resizebox{1.00\linewidth}{!}{
	\begin{tabular}{l| c| c c }
 
		\toprule
		Method & \makecell[c]{Input} & \makecell[c]{IoU} $\uparrow$ & \makecell[c]{mIoU} $\uparrow$
		  \\
		\midrule
Original GT  & $C$  & 29.5 & 9.4 \\
MigicDrive~\cite{gao2023magicdrive} & $C$ & 9.4 & 4.2 \\ 
Vista$^*$~\cite{gao2024vista} & $C$ & 14.3 & 5.1  \\  
 UniScene-C~\cite{li2024uniscene} & $C$ & 19.5 & 6.9\\ 
\rowcolor{gray!10} UniScenev2-C (Ours) & $C$ & 21.6 & 7.8 \\  
\midrule
Original GT & $L$ & 43.3 & 19.9 \\    
Open3D~\cite{zhou2018open3d} & $L$ & 6.5 & 3.3 \\
LiDAR-Diffusion~\cite{ran2024towards} & $L$ & 5.5 & 0.7 \\
UniScene-L~\cite{li2024uniscene} & $L$ & 30.8 & 10.0 
\\  
\rowcolor{gray!10} UniScenev2-L (Ours) & $L$ & 32.4 & 11.5
\\  
\bottomrule
\end{tabular}
}
\vspace{-0pt}
\caption{{Comparison about generation fidelity for the semantic occupancy prediction task} (Baseline as MonoScene~\cite{cao2022monoscene} and LMSCNet~\cite{roldao2020lmscnet}) on the Nuplan-Occ mini validation set.
The ``$C$'', ``$L$'', and ``$L^D$'' denote the camera, LiDAR, and depth projected from LiDAR, respectively.
}
\vspace{-0pt}
\label{tab_down1}
\end{center}
\end{table}

\begin{table}[!t]
\begin{center}
\scriptsize
\vspace{-0pt}
\renewcommand\tabcolsep{20.0pt}
	\centering
   \resizebox{1.00\linewidth}{!}{
	\begin{tabular}{l|  c c }
 
		\toprule
		Method &  \makecell[c]{NC} $\uparrow$ & \makecell[c]{DAC} $\uparrow$
		  \\
		\midrule
Original GT & 97.8 & 91.9 \\ \midrule
Vista$^*$~\cite{gao2024vista}  & 88.5 & 81.4  \\  
MigicDrive~\cite{gao2023magicdrive} & 91.6 & 85.7   \\ 
UniScene~\cite{li2024uniscene} & 93.1 & 86.1 \\
\rowcolor{gray!10} UniScenev2 (Ours) & 95.7 & 89.2
\\  
\bottomrule
\end{tabular}
}
\vspace{-0pt}
\caption{{Comparison about generation fidelity for the planning task} (baseline as UniAD~\cite{hu2023planning}) on the NAVSIM~\cite{dauner2024navsim}/Nuplan~\cite{Nuplan} test set.
}
\vspace{-0pt}
\label{tab_down2}
\end{center}
\end{table}

\noindent\textbf{Generation Fidelity Evaluation.}
We evaluate our model's ability to generate realistic driving scenarios using ground truth occupancy conditions. Unlike existing works~\cite{gao2023magicdrive,swerdlow2024street} that generate only RGB images, our approach produces multiple data modalities, enabling a comprehensive evaluation for downstream multi-modal tasks. As shown in Table~\ref{tab_down1}, \ours\ outperforms other methods in both camera-based and LiDAR-based semantic occupancy prediction. Furthermore, we evaluate generation fidelity for the navigation planning task on the NAVSIM~\cite{dauner2024navsim}/Nuplan test set, where our method surpasses alternatives on the no at-fault collisions (NC) and drivable area compliance (DAC) metrics. These results demonstrate the high quality and potential of our synthetic data for diverse multi-modal applications.

\begin{table}[!t]
\rcap
\vspace{-0pt}
\centering
\renewcommand\tabcolsep{2.0pt}
\centering
\resizebox{0.99\linewidth}{!}{
\centering
\begin{tabular}{l|ccc}
\toprule
Generated Training Data Scale & Data Size (Frames) & NC $\uparrow$ & DAC $\uparrow$ \\ \midrule
10\% Nuplan-Occ (Similar to UniScene) & $\sim$360K & 93.4 & 86.5 \\
50\% Nuplan-Occ  & $\sim$1.8M & 94.6 & 88.1 \\
100\% Nuplan-Occ (UniScenev2) & $\sim$3.6M & \textbf{95.7} & \textbf{89.2} \\
\bottomrule
\end{tabular}
}
\vspace{5pt}
\caption{
\re{{Generation Data Scaling Evaluation.} Impact of generated data volume on downstream planning metrics using the NAVSIM~\cite{dauner2024navsim}/Nuplan~\cite{Nuplan} test set.}}
\label{re_scaling_planning}
\vspace{-0pt}
\end{table}

\re{To further validate the effectiveness of our generated realistic driving scenarios with ground truth occupancy conditions, we evaluate downstream planning performance across different generation data volumes as shown in Table~\ref{re_scaling_planning}. The results demonstrate that the planning metrics (NC and DAC) compellingly improve as the volume of high-quality generation data increases. Notably, reducing the generated training data to 10\% of Nuplan-Occ (a scale comparable to previous benchmarks) degrades the NC metric back to 93.4\%, confirming that the scale of the generated data is a direct factor in overcoming previous downstream performance bottlenecks.}

\begin{figure}[!t]
    \centering
    \vspace{-0pt}
    \includegraphics[width=1.0\linewidth]{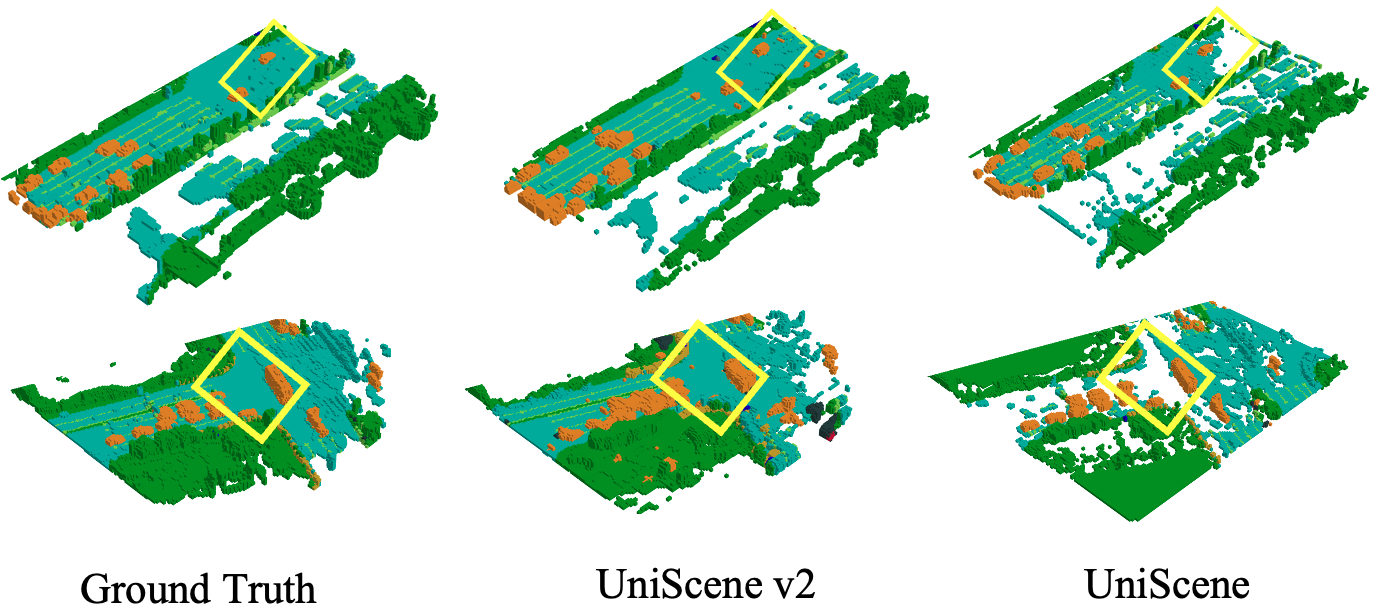}
    \vspace{-0pt}
    \caption{{Qualitative evaluation} for occupancy generation. 
    Our method generates more complete and accurate scene layouts.}
    \label{occ_compare}
    \vspace{-0pt}
\end{figure}

\begin{figure}[!t]
    \centering
    \includegraphics[width=0.99\linewidth]{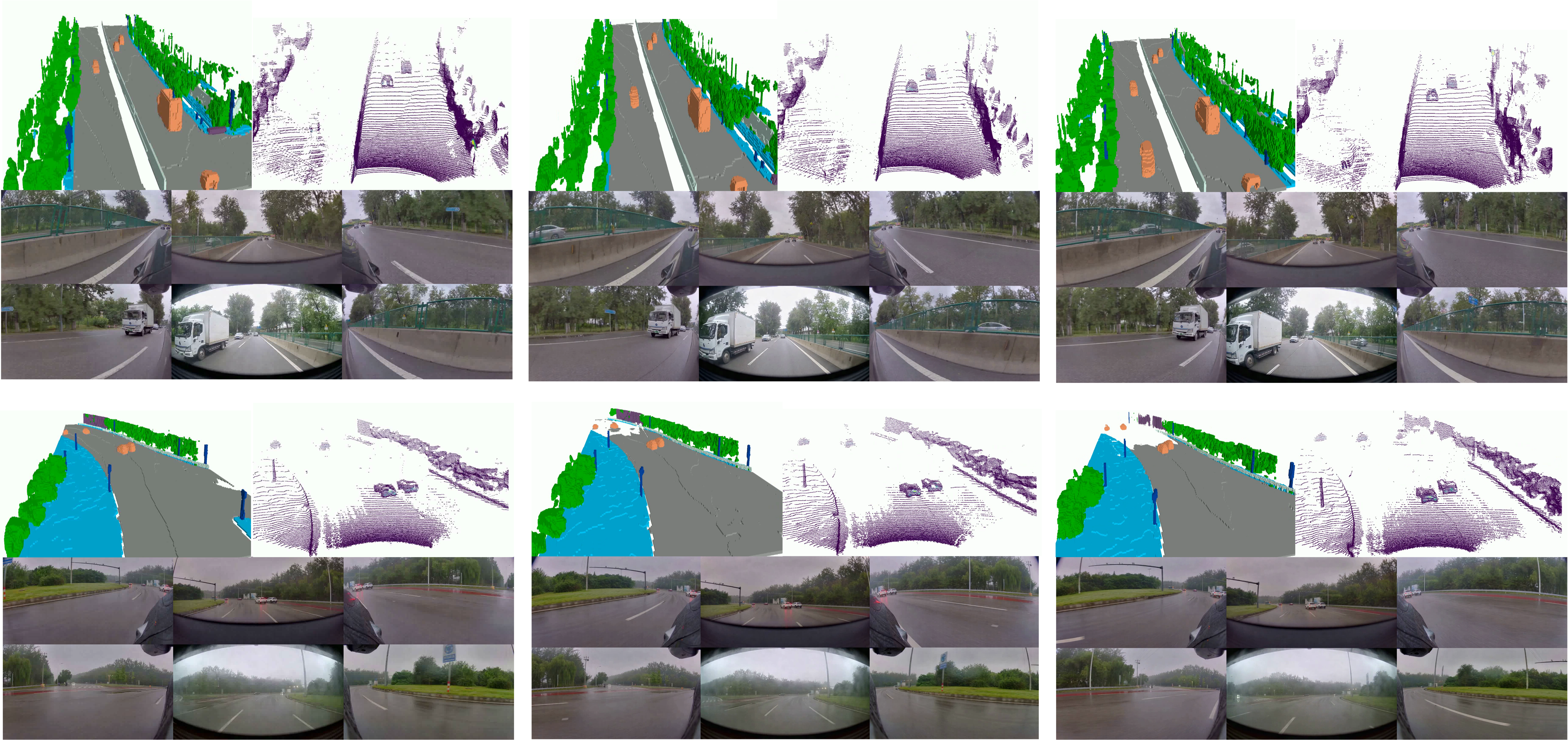}
    \vspace{-0pt}
    \caption{Generalizable generation on the in-house collected datasets with different sensor configurations of 6 fisheye cameras and a front LiDAR sensor.   
    }
    \label{fig_generalizable}
    \vspace{-0pt}
\end{figure}

\noindent\textbf{Generalizable Generation.}
Figure~\ref{fig_generalizable} presents generalizable generation results. We evaluate \ours\ on the in-house collected datasets with totally different sensor configurations (\ie, 6 fisheye cameras and a front LiDAR sensor). 
As we can see, our method demonstrates strong generalization capabilities on distinct settings, producing high-quality generation results of 3D occupancy, multi-view video, and LiDAR data.

\subsection{Ablation Studies}\label{ablation}

\begin{table}[!t]
\vspace{-0pt}
\centering
\renewcommand\tabcolsep{11.0pt}
\resizebox{1.00\columnwidth}{!}{
\begin{tabular}{l|ccc}
\toprule
        Method & mIoU $\uparrow$  & F3D $\downarrow$ & MMD $\downarrow$ \\
         \midrule 
    \rowcolor{gray!10}Ours & 32.22&48.24&0.784
    \\ \midrule
    w/o. VAE 3D Axial Attention &  21.42 & 140.23& 10.79\\
    w/o. DiT Temporal Attention &  19.32& 170.34 & 11.48 \\
    w/o. DiT Spatial Attention &  15.67 & 240.73 & 17.21 \\
    w/o. BEV Condition & 17.13 & 178.24 & 13. 67 \\
    \bottomrule
    \end{tabular}%
}
\vspace{-0pt}
\caption{{Ablation} for designs in the occupancy generation model on the Nuplan-mini validation set.
} 
\label{table_ab_occ_1}
\vspace{-0pt}
\end{table}

\noindent\textbf{Effect of Designs in Occupancy Generation Model.}\label{abl_occ} 
We conduct ablation studies to evaluate the contribution of key components in our occupancy generation model, as summarized in Tab.~\ref{table_ab_occ_1}. Incorporating temporal information into the occupancy VAE decoder through 3D axial attention significantly enhances the fidelity of occupancy sequence generation, reflected by a 33.52\% improvement in mIoU. Both the temporal and spatial attention layers in the occupancy DiT substantially improve generation quality, increasing the F3D metric by 40.04\% and 51.37\%, respectively.

\begin{table}[!t]
\vspace{-0pt}
\centering
\renewcommand\tabcolsep{5.0pt}
\resizebox{1.00\columnwidth}{!}{
\begin{tabular}{l|cccc}
\toprule
Method & Gaussian Scale & FID$\downarrow$ & FVD$\downarrow$ \\ \midrule
\rowcolor{gray!10}Ours & 0.01 & 8.32 & 63.29  \\ \midrule
w/o. Sparse Rendered Semantic Map & - & 12.27 & 110.79  \\
w/o. Sparse Rendered Depth Map   & -  & 12.05 & 108.21  \\ 
w/o. Unscented Transform Calibration & - & 9.26  & 72.91 \\  
\midrule
\multirow{3}{*}{w/. Rendered Maps}
 & 0.002 & 8.76 & 74.28 \\
 & 0.01  & 8.32 & 63.29  \\
 & 0.04  & 9.21 & 78.69\\  

 \bottomrule
\end{tabular}
}
\vspace{-0pt}
\caption{{Ablation} for designs in the video generation model on the Nuplan-mini validation set.
} 
\label{table_ar_video}
\vspace{-0pt}
\end{table}

\noindent\textbf{Effect of Designs in Video Generation Model.}
We conduct ablation studies to evaluate the components of our video generation model, as summarized in Table~\ref{table_ar_video}. The results demonstrate that occupancy-based semantic and geometric sparse rendering maps are more effective for improving video quality than other conditioning inputs. Furthermore, a Gaussian scale of 0.01 yields the best performance, achieving FID and FVD scores of 8.32 and 63.29 with ground truth occupancy.\looseness=-1

\begin{table}[!t]
\vspace{-0pt}
\centering
\renewcommand\tabcolsep{1.0pt}
\resizebox{1.00\columnwidth}{!}{
\begin{tabular}{l|cccc}
\toprule
Method & MMD ($10^{-5}$) $\downarrow$ &JSD $\downarrow$ & Time (s)$\downarrow$ & Memory (GB)$\downarrow$   \\ \midrule
\rowcolor{gray!10}Ours  & 0.457 & 0.028 & 0.36 & 12.76 \\ \midrule
w/o. Sensor-specific Embedding  & 0.783 & 0.032 & 0.32 & 12.76  \\ 
w/o. Plücker Embedding & 0.908 & 0.034 & 0.34 & 12.16\\ 
w/o. Histogram Embedding  & 0.997 & 0.036 & 0.34 & 11.17 \\
w/o. Smoothness Regularization & 0.694 & 0.030 & 0.36 & 12.76 \\
\bottomrule
\end{tabular}
}
\vspace{-0pt}
\caption{
{Ablation} for designs in the LiDAR generation model on the Nuplan-mini validation set.
} 
\label{table_ar_lidar}
\vspace{-0pt}
\end{table}

\noindent\textbf{Effect of Designs in LiDAR Generation Model.}
Ablation studies on the key components of our LiDAR generation model is summarized in Table~\ref{table_ar_lidar}. The complete model achieves the best performance, with an MMD of \(0.457 \times 10^{-5}\) and a JSD of 0.028. Removing the sensor-specific embedding results in a significant performance drop, increasing MMD by 41.63\%. Similarly, omitting Plücker embedding or histogram embedding degrades MMD by 49.67\% and 54.16\%, respectively, confirming their importance in representing LiDAR characteristics. The smoothness regularization also contributes to model stability, with its removal increasing MMD by 34.14\%. All variants have comparable inference time and memory usage, indicating that the performance gains are not achieved at the cost of efficiency.

\section{Conclusion} 
In this paper, we presented \ours, a scalable framework for unified occupancy-centric driving scene generation. The proposed method synthesizes high-quality semantic occupancy grids, multi-view videos, and LiDAR point clouds in a unified pipeline. By introducing a spatio-temporal disentangled architecture and an effective data filtering strategy, our approach supports robust spatial expansion and temporal forecasting, enabling large-scale 4D occupancy generation.
To bridge modality gaps, we introduced two key technical innovations: a Gaussian Splatting-based sparse point map rendering method for video generation, and a sensor-specific embedding strategy for realistic LiDAR simulation. Furthermore, we contributed Nuplan-Occ, the largest semantic occupancy dataset to date, to facilitate scalable training and evaluation.
Extensive experiments validate that \ours\ outperforms existing state-of-the-art methods across occupancy, video, and LiDAR generation tasks. The framework also demonstrates strong potential in enhancing downstream applications, underscoring its practical value for autonomous driving research.

\section*{Acknowledgments}
This work was supported by Grants of NSFC 62302246, ZJNSFC LQ23F010008, Ningbo 2023Z237 \& 2024Z284 \& 2024Z28\& 2023CX050011 \& 2025Z038 \& 2025Z059, and supported by High Performance Computing Center at Eastern Institute of Technology and Ningbo Institute of Digital Twin. 


\bibliographystyle{IEEEtran}
 
\bibliography{egbib}

\vspace{10pt}
\begin{IEEEbiography}[{\includegraphics[width=1in,height=1.25in,clip,keepaspectratio]{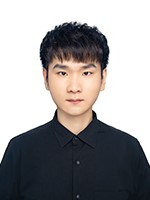}}]{Bohan Li} received the B.E. degree from the School of Control Engineering, Northeastern University (NEU),
Shenyang, China, in 2019. He received the M.E. degree from the School of Control Science and Engineering, South China University of Technology
(SCUT), Guangzhou, China, in 2022.

He is currently pursuing the Ph.D. degree in Shanghai Jiao Tong University (SJTU) and Eastern Institute of Technology (EIT). His research interests include multi-modality content generation, 3D visual perception, autonomous driving, and robotics.
\end{IEEEbiography}
\vspace{-2em}

\begin{IEEEbiography}[{\includegraphics[width=1in,height=1.25in,clip,keepaspectratio]{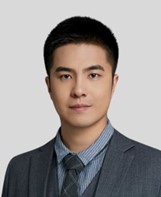}}]{Xin Jin} has been a tenure track Assistant Professor with the Eastern Institute of Technology (EIT), Ningbo, China. He is also a Researcher at the Ningbo Institute of Digital Twin. He received his Ph.D. degree in Electronic Engineering and Information Science from the University of Science and Technology of China (USTC). His research interests include computer vision, intelligent media computing, and deep learning. He has over 10 granted patent applications, around 40 publications, and over 3,500 Google citations. He is an IEEE member, and reviewer of IEEE Transactions on Image Processing (TIP), IEEE Transactions on Multimedia (TMM), and IEEE Transactions on Circuits and Systems for Video Technology (TCSVT).
\end{IEEEbiography}
\vspace{-2em}

\begin{IEEEbiography}[{\includegraphics[width=1in,height=1.25in,clip,keepaspectratio]{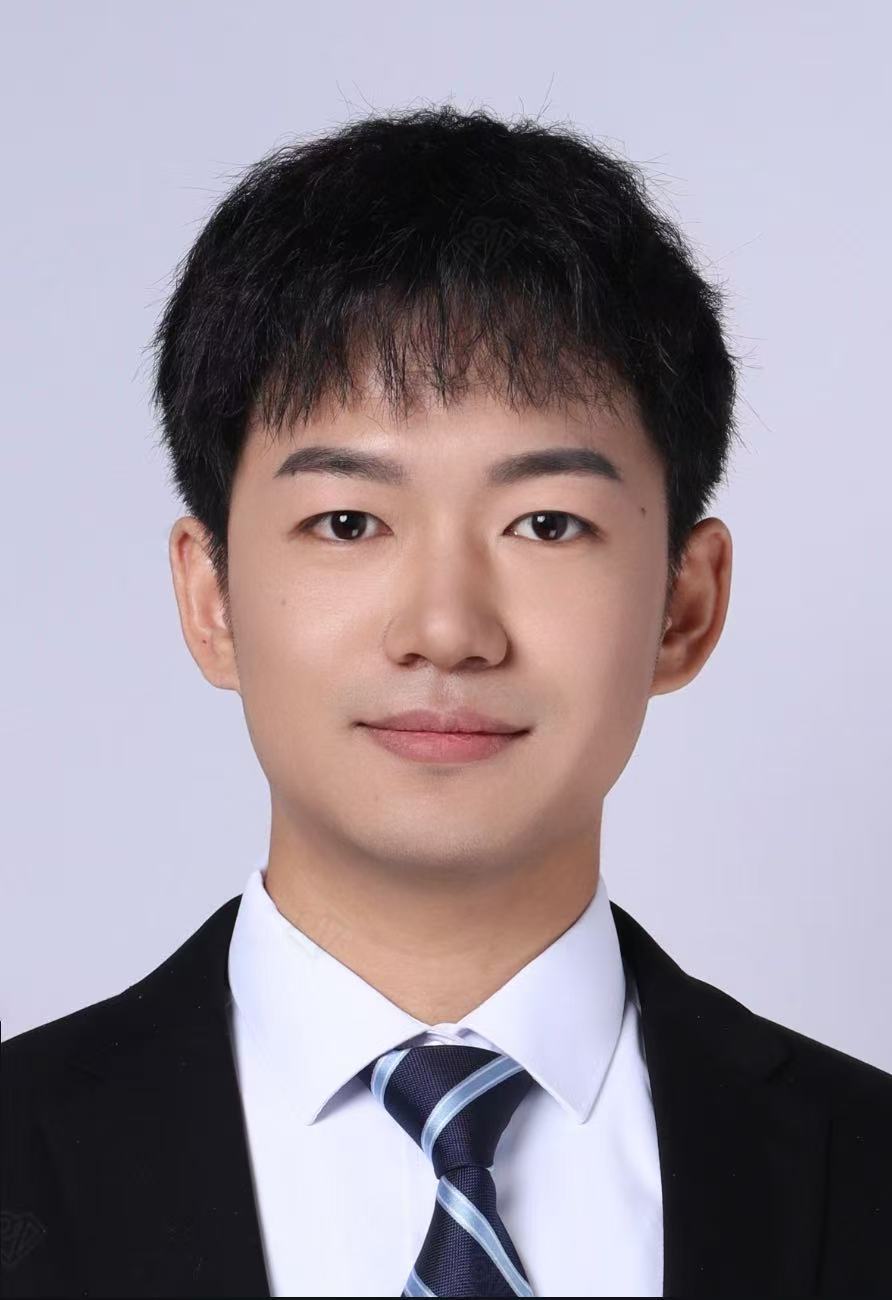}}]{Hu Zhu}received the B.E. degree from the School of Automation Engineering, Xi‘an Jiaotong University(XJTU),
Xi’an, China, in 2020. He received the M.E. degree from the School of Engineering , Southern University of Science and Technology
(SUSTech), Shenzhen, China, in 2023.
He is currently pursuing the Ph.D. degree in  The Hong Kong Polytechnic University(PolyU) and Eastern Institute of Technology (EIT). His research interests include 3D content generation.
\end{IEEEbiography}
\vspace{-2em}

\begin{IEEEbiography}[{\includegraphics[width=1in,height=1.25in,clip,keepaspectratio]{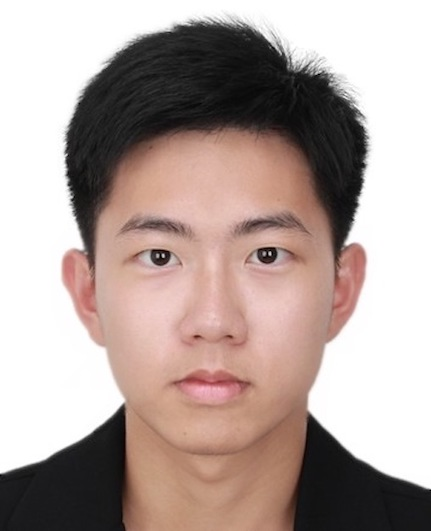}}]{Hongsi Liu}received the B.S. degree in Electronic Information Science and Technology from Sun Yat-sen University (SYSU), Shenzhen, China, in 2022. He is currently pursuing the Ph.D. degree in University of Science and Technology of China (USTC) and Eastern Institute of Technology (EIT). His research interests include autonomous driving, robotics, 3D understanding, generation, and reconstruction.
\end{IEEEbiography}
\vspace{-2em}

\begin{IEEEbiography}[{\includegraphics[width=1in,height=1.25in,clip,keepaspectratio]{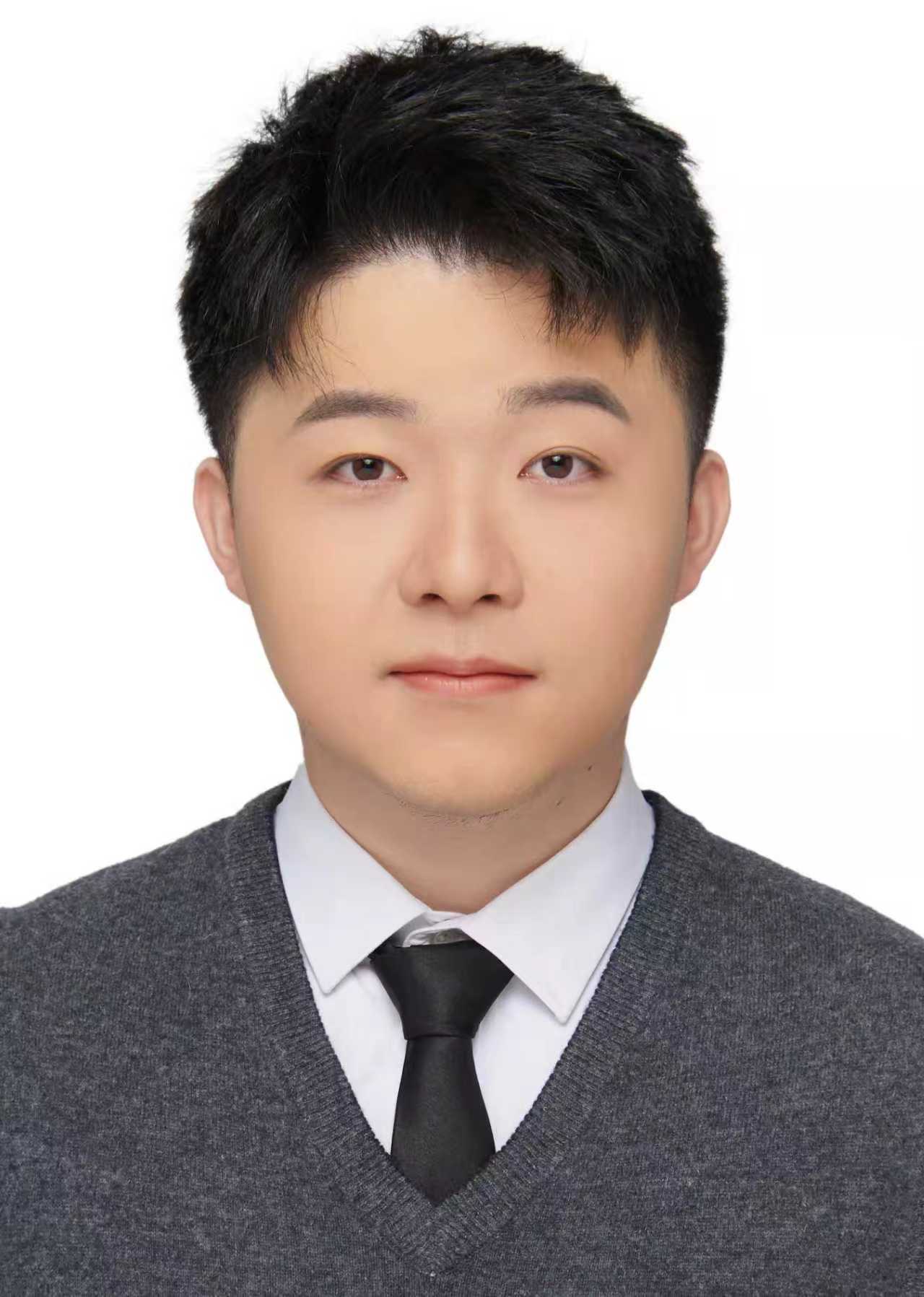}}]{Ruikai Li} received his B.S. degree from the School of Software, Beijing University of Technology (BJUT), under a joint program with Beihang University (BHU) in 2022. He is currently pursuing the Ph.D. degree at the School of Transportation Science and Engineering, Beihang University. His research interests include computer vision and model compression for intelligent transportation systems.
\end{IEEEbiography}
\vspace{-2em}

\begin{IEEEbiography}[{\includegraphics[width=1in,height=1.25in,clip,keepaspectratio]{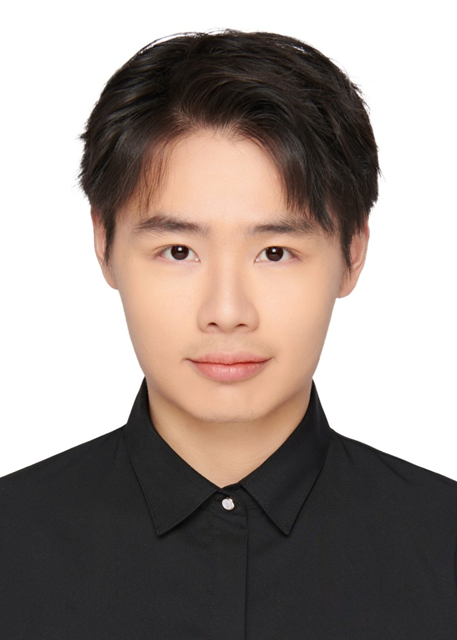}}]{Jiazhe Guo}received the B.E. degree from the College of Control Science and Engineering, Zhejiang University (ZJU), Hangzhou, China, in 2023. He is currently pursuing the M.E. degree in Shenzhen International Graduate School, Tsinghua University (THU), Shenzhen, China. His research interests include autonomous driving and generative models. 
He is a reviewer of computer vision conferences, including CVPR, ECCV, ICCV, AAAI, and ICLR.
\end{IEEEbiography}
\vspace{-2em}

\begin{IEEEbiography}[{\includegraphics[width=1in,height=1.25in,clip,keepaspectratio]{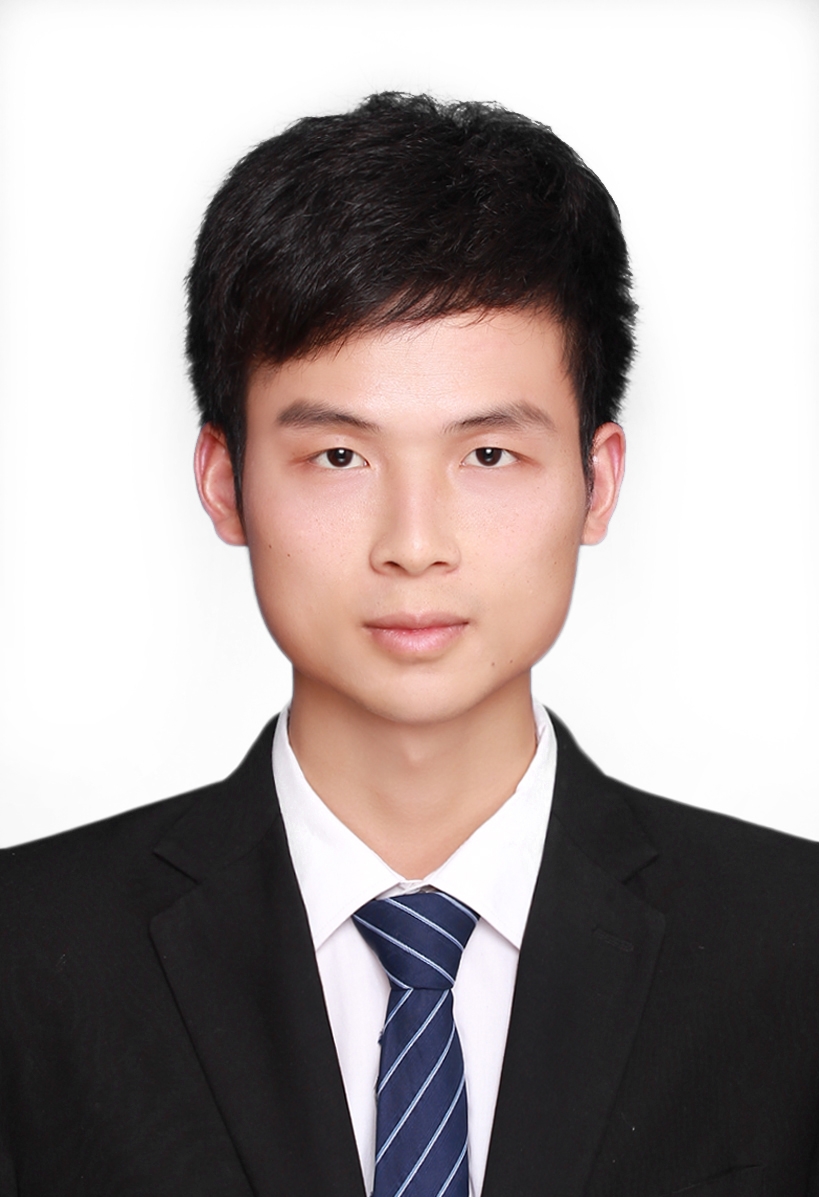}}]{Kaiwen Cai} received the B.S. and M.S. degrees in 2017 and 2020, respectively, from Nanjing University of Aeronautics and Astronautics (NUAA), Nanjing, China, and the Ph.D. degree from the University of Liverpool (UOL), U.K., in 2024. He is currently a researcher at LiAuto Corporation, where his work focuses on computer vision, autonomous driving, generative models.
\end{IEEEbiography}
\vspace{-2em}

\begin{IEEEbiography}[{\includegraphics[width=1in,height=1.25in,clip,keepaspectratio]{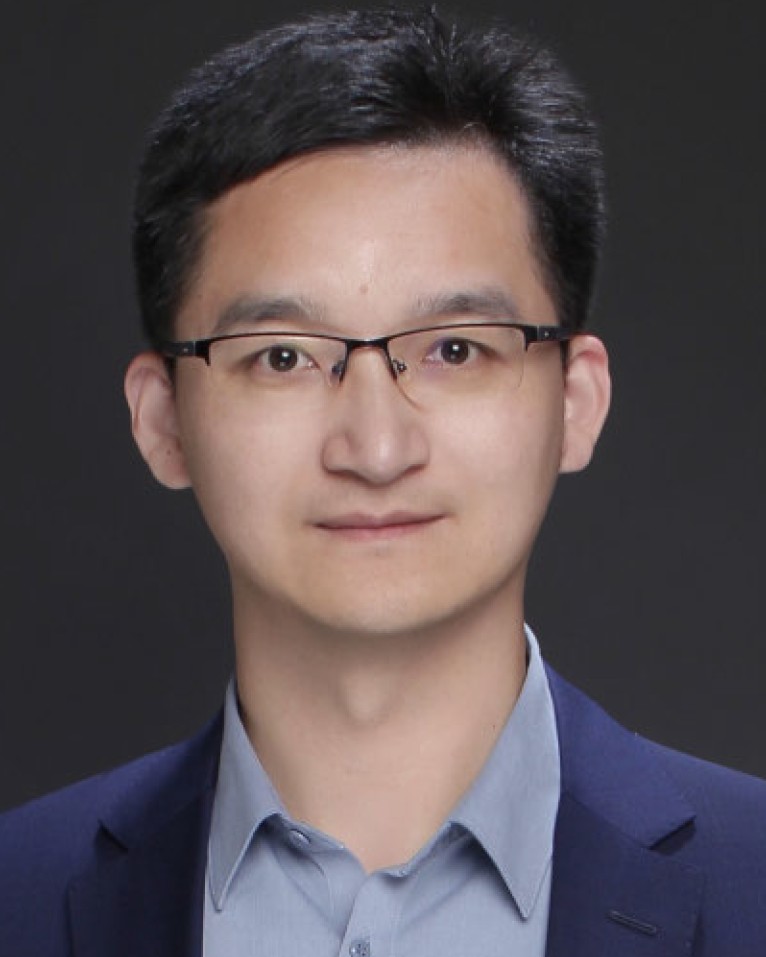}}]{Chao Ma} (Member, IEEE) received the Ph.D. degree from Shanghai Jiao Tong University, Shanghai, China, in 2016. He was sponsored by the China Scholarship Council as a Visiting Ph.D. Student at the University of California at Merced, Merced, CA, USA, from Fall 2013 to Fall 2015. He was a Research Associate with the School of Computer Science, The University of Adelaide, Adelaide, SA, Australia, from 2016 to 2018. He is currently a Professor at Shanghai Jiao Tong University. His research interests include computer vision and machine learning.
\end{IEEEbiography}
\vspace{-2em}

\begin{IEEEbiography}[{\includegraphics[width=1in,height=1.25in,clip,keepaspectratio]{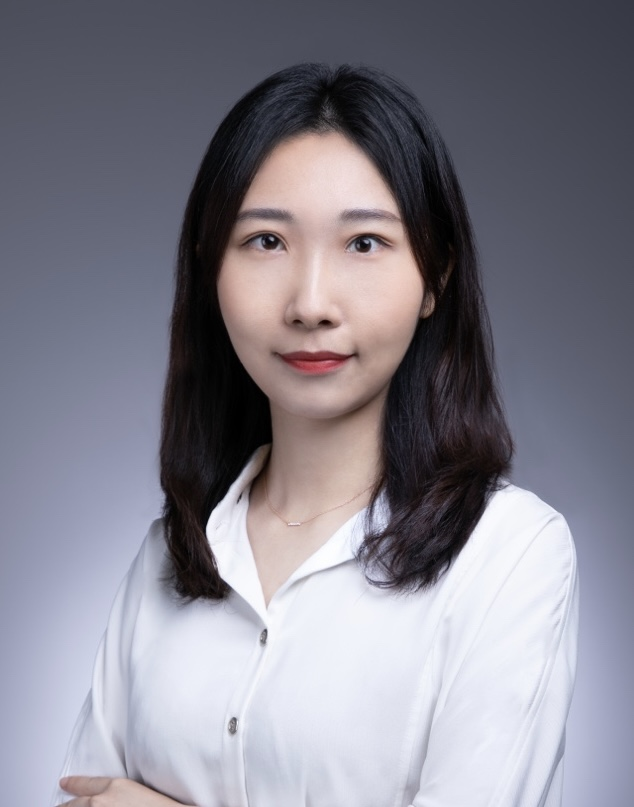}}]{Yueming Jin} received Ph.D. degree in the Department of Computer Science and Engineering, The Chinese University of Hong Kong. She is currently an Assistant Professor at Department of Biomedical Engineering and Department of Electrical and Computer Engineering at National University of Singapore. Her research interests include artificial intelligence and its applications on healthcare, with an emphasized application to medical image computing and robotic surgical data science.
\end{IEEEbiography}
\vspace{-2em}

\begin{IEEEbiography}[{\includegraphics[width=1in,height=1.25in,clip,keepaspectratio]{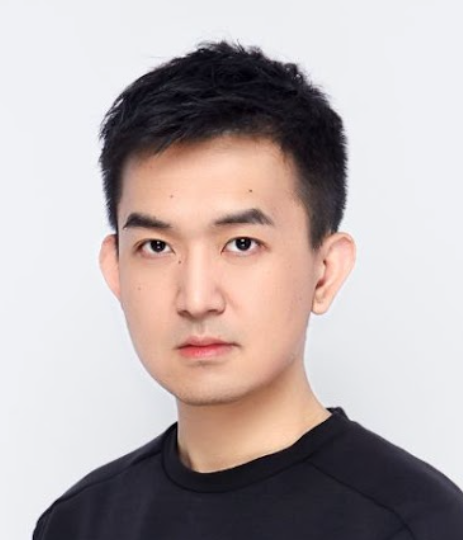}}]{Hao Zhao} received the B.E. degree and the Ph.D. degree both from the EE department of Tsinghua University, Beijing, China. He is currently an Assistant Professor with the Institute for AI Industry Research (AIR), Tsinghua University. He was a research scientist at Intel Labs China and a joint postdoc affiliated to Peking University. His research interests cover various computer vision topics related to robotics, especially 3D scene understanding. Photograph not available at the time of publication.
\end{IEEEbiography}
\vspace{-2em}

\begin{IEEEbiography}[{\includegraphics[width=1in,height=1.25in,clip,keepaspectratio]{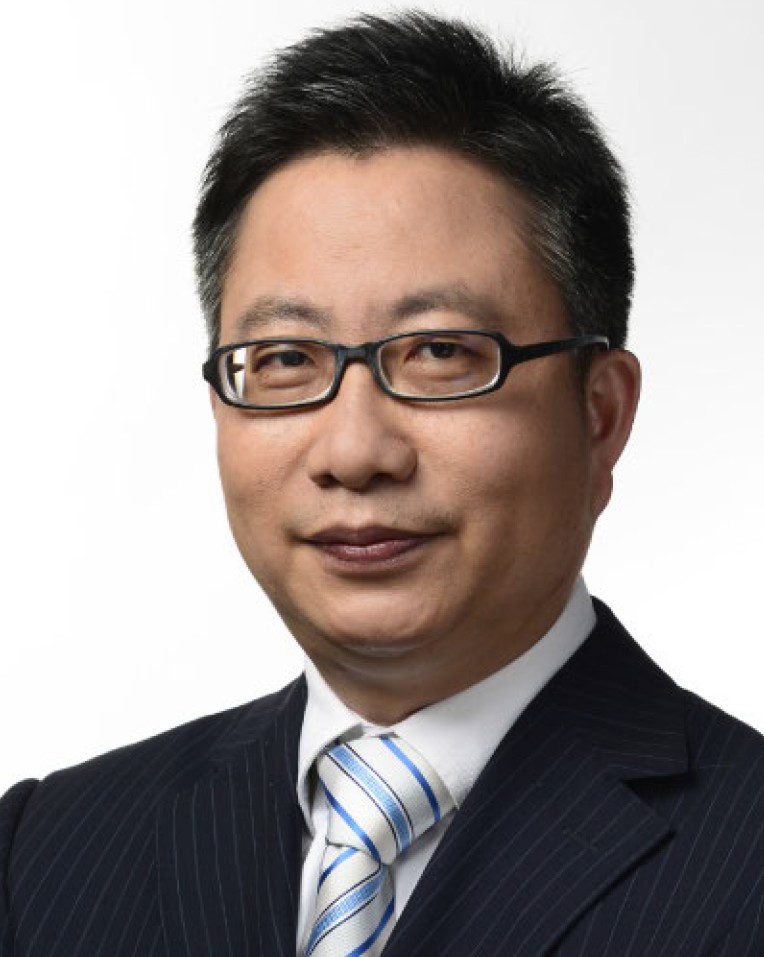}}]{Xiaokang Yang} (Fellow, IEEE) received the B.S. degree from Xiamen University, Xiamen, China, in 1994, the M.S. degree from the Chinese Academy of Sciences, Shanghai, China, in 1997, and the Ph.D. degree from Shanghai Jiao Tong University, Shanghai, in 2000. From September 2000 to March 2002, he worked as a Research Fellow with the Centre for Signal Processing, Nanyang Technological University, Singapore. From April 2002 to October 2004, he was a Research Scientist at the Institute for Infocomm Research (I2R), Singapore. From August 2007 to July 2008, he visited the Institute for Computer Science, University of Freiburg, Breisgau, Germany, as an Alexander von Humboldt Research Fellow. He is currently a Distinguished Professor at the School of Electronic Information and Electrical Engineering, Shanghai Jiao Tong University. He has published over 200 refereed articles and has filed 60 patents. His research interests include image processing and communication, computer vision, and machine learning. Dr. Yang received the 2018 Best Paper Award of IEEE TRANSACTIONS ON MULTIMEDIA. He is an Associate Editor of IEEE TRANSACTIONS ON MULTIMEDIA and a Senior Associate Editor of IEEE SIGNAL PROCESSING LETTERS.
 \end{IEEEbiography}
\vspace{-2em}

\begin{IEEEbiography}[{\includegraphics[width=1in,height=1.25in,clip,keepaspectratio]{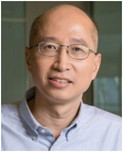}}]{Wenjun Zeng} (Fellow, IEEE) received the B.E. degree from Tsinghua University, Beijing, China,
in 1990, the M.S. degree from the University of Notre Dame, Notre Dame, IN, USA, in 1993, and the Ph.D. degree from Princeton University, Princeton, NJ, USA, in 1997. He has been a Chair Professor and the Vice President for Research at the Eastern Institute for Advanced Study (EIAS) / Eastern Institute of Technology (EIT), Ningbo, China, since October 2021. He is also the founding Executive Director of the Ningbo Institute of Digital Twin. He was a Sr. Principal Research Manager and a member of the Senior Leadership Team at Microsoft Research Asia, Beijing, from 2014 to 2021, where he led the video analytics research empowering the Microsoft Cognitive Services, Azure Media Analytics Services, Office, and Windows Machine Learning. He was with University of Missouri, Columbia, MO, USA from 2003 to 2016, most recently as a Full Professor. Prior to that, he had worked for PacketVideo Corp., Sharp Labs of America, Bell Labs, and Panasonic Technology. He has contributed significantly to the development of international standards (ISO MPEG, JPEG2000, and OMA).
Dr. Zeng is on the Editorial Board of the International Journal of Computer Vision. He was an Associate Editor-in-Chief of the IEEE Multimedia Magazine and an Associate Editor of the IEEE TRANSACTIONS ON CIRCUITS AND SYSTEMS FOR VIDEO TECHNOLOGY, IEEE TRANSACTIONS ON INFORMATION FORENSICS AND SECURITY, and IEEE TRANSACTIONS
ON MULTIMEDIA (TMM). He was on the Steering Committee of IEEE TRANSACTIONS ON MOBILE COMPUTING and IEEE TMM. He served as the Steering Committee Chair of IEEE ICME in 2010 and 2011, and has served as the General Chair or TPC Chair for several IEEE conferences (\eg, ICME’2018, ICIP’2017). He was the recipient of several best paper awards.
\end{IEEEbiography}
\vspace{-2em}

\vfill

\end{document}